%% file: icml2026.tex
\documentclass{article}

\usepackage{microtype}
\usepackage{graphicx}
\usepackage{subcaption}
\usepackage{booktabs} 

\usepackage{hyperref}

\usepackage[preprint]{icml2026}

\usepackage{graphicx} 

\newcommand*{\dashline}{
    \rotatebox[origin=c]{90}{
      \tiny{- -}
    }
}
\usepackage{enumitem}
\usepackage{amsmath}
\usepackage{amssymb}
\usepackage{mathtools}
\usepackage{amsthm}
\usepackage{graphicx}
\usepackage{booktabs}
\usepackage{subcaption}
\usepackage{colortbl}
\usepackage{makecell}
\usepackage{caption}
\usepackage{multirow}
\usepackage{listings}
\usepackage{cuted}
\usepackage{multicol}
\usepackage{titling}
\usepackage{manyfoot}
\usepackage{verbatim}
\usepackage{xspace}
\usepackage{marginnote}
\usepackage{xcolor}
\usepackage[capitalize,noabbrev]{cleveref}
\definecolor{basecolor}{RGB}{220,220,220}   
\definecolor{sftcolor}{RGB}{180,205,230}    
\definecolor{grpocolor}{RGB}{200,230,200}   

\definecolor{humanlmcolor}{RGB}{245,200,200} 

\definecolor{improvecolor}{RGB}{255,230,200} 

\theoremstyle{plain}

\theoremstyle{definition}

\theoremstyle{remark}

\usepackage{siunitx}
\usepackage[textsize=tiny]{todonotes}

\newcommand{\ie}{\textit{i.e., }}
\newcommand{\eg}{\textit{e.g., }}
\newcommand{\vs}{\textit{v.s. }}

\newcommand{\etc}{\textit{etc.}}

\newcommand{\cf}{\textit{cf. }}
\newcommand{\aka}{\textit{aka. }}

\newcommand{\reddit}{\texttt{\benchmarkvanilla-Opinion}}
\newcommand{\medium}{\texttt{\benchmarkvanilla-Politics}}
\newcommand{\youtube}{\texttt{\benchmarkvanilla-News}}
\newcommand{\amazon}{\texttt{\benchmarkvanilla-Book}}
\newcommand{\wildchat}{\texttt{\benchmarkvanilla-Chat}}
\newcommand{\enron}{\texttt{\benchmarkvanilla-Email}}

\newcommand{\redditabbr}{\texttt{Opinion}}
\newcommand{\mediumabbr}{\texttt{Politics}}
\newcommand{\youtubeabbr}{\texttt{News}}
\newcommand{\amazonabbr}{\texttt{Book}}
\newcommand{\wildchatabbr}{\texttt{Chat}}
\newcommand{\enronabbr}{\texttt{Email}}

\newcommand{\claude}{\texttt{claude-4.5-haiku}}

\newcommand{\baseline}{{Qwen3-8b}}
\newcommand{\baselinethink}{{Qwen3-8b-think}}

\newcommand{\simulator}{user simulator}
\newcommand{\base}{Qwen3-8b}
\newcommand{\name}{\textsc{HumanLM}}
\newcommand{\benchmark}{\textsc{Humanual}} 
\newcommand{\benchmarkvanilla}{Humanual} 
\newcommand{\dimension}{state dimension}
\newcommand{\state}{latent state}
\newcommand{\State}{Latent State}
\newcommand{\benchmarkimprov}{16.3\%}

\newcommand{\humanualusers}{26k} 
\newcommand{\humanualposts}{66k} 
\newcommand{\humanualresponses}{216k} 

\newcommand{\nuser}{111}

\newcommand{\humanlmsim}{6.5}

\newcommand{\basewinrate}{30.6\%} 
\newcommand{\grpowinrate}{27.9\%} 
\newcommand{\winrate}{41.4\%}     
\newcommand{\winrategapbase}{-10.8\%} 
\newcommand{\winrategapgrpo}{-13.5\%}
\newcommand{\humanlmabovebarsim}{55.9\%} 
\newcommand{\humanlmabovebarlike}{76.6\%} 
\newcommand{\baseabovebarsim}{45.0\%} 
\newcommand{\baseabovebarlike}{72.1\%}

\newcommand{\titlename}{\name: Simulating Users with State Alignment Beats Response Imitation}

\usepackage{tikz}
\newcommand*\circled[1]{\tikz[baseline=(char.base)]{
            \node[shape=circle,draw,inner sep=0.5pt] (char) {#1};}}

\newcommand{\xhdrd}[1]{{\noindent\bfseries #1}.}
\newcommand{\xhdr}[1]{{\noindent\bfseries #1}}
\icmltitlerunning{\titlename{}}

\begin{document}
\onecolumn 

  \icmltitle{\titlename}
  \icmlsetsymbol{equal}{*}

  \begin{icmlauthorlist}
    \icmlauthor{Shirley Wu}{equal,stanford}
    \icmlauthor{Evelyn Choi}{equal,stanford}
    \icmlauthor{Arpandeep Khatua}{equal,stanford}
    \icmlauthor{Zhanghan Wang}{nyu}
    \icmlauthor{Joy He-Yueya}{stanford}
    \icmlauthor{Tharindu Cyril Weerasooriya}{accenture}
    \icmlauthor{Wei Wei}{accenture}
    \icmlauthor{Diyi Yang}{stanford}
    \icmlauthor{Jure Leskovec$^{**}$}{stanford}
    \icmlauthor{James Zou$^{**}$}{stanford}\\
    {\small{$^*$\textit{Equal Contribution}\quad \quad$^{**}$\textit{Equal Senior Supervision}}}\\
    \vspace{3pt}
    \url{http://humanlm.stanford.edu/}
  \end{icmlauthorlist}

  \icmlaffiliation{stanford}{Stanford University}
  \icmlaffiliation{nyu}{New York University}
  \icmlaffiliation{accenture}{Accenture}

  \icmlcorrespondingauthor{}{\texttt{\small\{shirwu,jure,jamesz\}@cs.stanford.edu}}

  \icmlkeywords{Machine Learning, ICML}

  \vskip 0.3in
\printAffiliationsAndNotice{}

\begin{figure}[H] 
    \centering
    \vspace{-25pt}
    \includegraphics[width=1.0\linewidth]{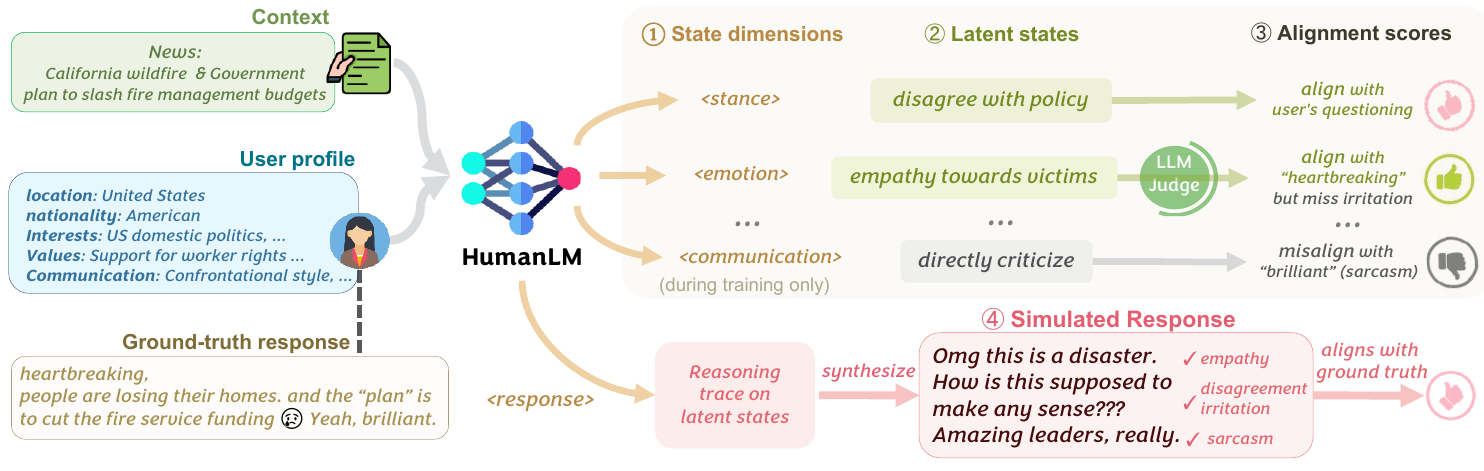}
    \vspace{-20pt}
    \captionof{figure}{
    \name{} generates responses that capture the key points of real user responses.
    Given an input context (\eg a news post) and a user profile, the model prioritizes alignment along a few psychologically grounded \circled{\small 1}  \dimension{}s (\eg stance, emotion), that lead to how users respond. 
    For each \dimension{}, the model generates the corresponding \circled{\small 2} \state{} (\eg ``empathy toward victims''), \circled{\small 3} scored by an LLM judge for consistency with the ground-truth response.
    During reinforcement learning, the model maximizes alignment scores on \state{}s to accurately reflect real users, in addition to directly improving the responses.
    When generating responses, the model generates reasoning traces with aligned \state{}s to synthesize \circled{\small 4} accurate responses.
    }
    \label{fig:overview}
\end{figure}

\begin{multicols}{2} 
\input{chapters/1_abstract}
\input{chapters/2_introduction}

\input{chapters/4_method}

\input{chapters/5_experiments}
\input{chapters/3_related_works}
\input{chapters/6_conclusion}

\bibliography{reference}
\bibliographystyle{icml2026}

\end{multicols}  
\newpage
\onecolumn
\appendix
\input{chapters/7_appendix}

\end{document}

%% file: chapters/1_abstract.tex

\begin{abstract}
    Large Language Models (LLMs) are increasingly used to simulate how specific users respond to any context, enabling more user-centric applications that rely on user feedback.
    However, existing \simulator{}s mostly imitate surface-level patterns and language styles, which fails to reflect the underlying state of real users (\eg beliefs, emotions).
    To address these limitations, we propose a novel training framework, \name{}, which builds \simulator{}s that accurately reflect real users.
    Our key insight is, in addition to generating responses, we generate natural-language \textit{\state{}s} that align with the ground truth responses 
    through reinforcement learning.
    These \state{}s correspond to a set of \textit{\dimension{}}s which psychologically lead to 
    how real users 
    respond. \name{} further synthesizes these aligned \state{}s into responses that accurately represent real users. For extensive evaluation, we 
    develop \benchmark{}, a comprehensive benchmark on simulating real users based on public data. \benchmark{} consists of six large-scale datasets with \humanualusers{} users and \humanualresponses{} responses in total. It spans diverse tasks such as generating user responses to daily life issues, political blogs, and chat sessions with LLM assistants.
    Across the datasets, \name{} significantly outperforms the best alternative approaches by an average relative improvement of \benchmarkimprov{} on alignment score from an LLM judge. In a real-time simulation study with \nuser{} participants, \name{} achieves the highest scores on similarity with real user responses and humanlikeness. 
\end{abstract}

%% file: chapters/2_introduction.tex
\vspace{-15pt}

\section{Introduction}
\begin{figure*}[t]
    \centering
    \includegraphics[width=0.99\linewidth]{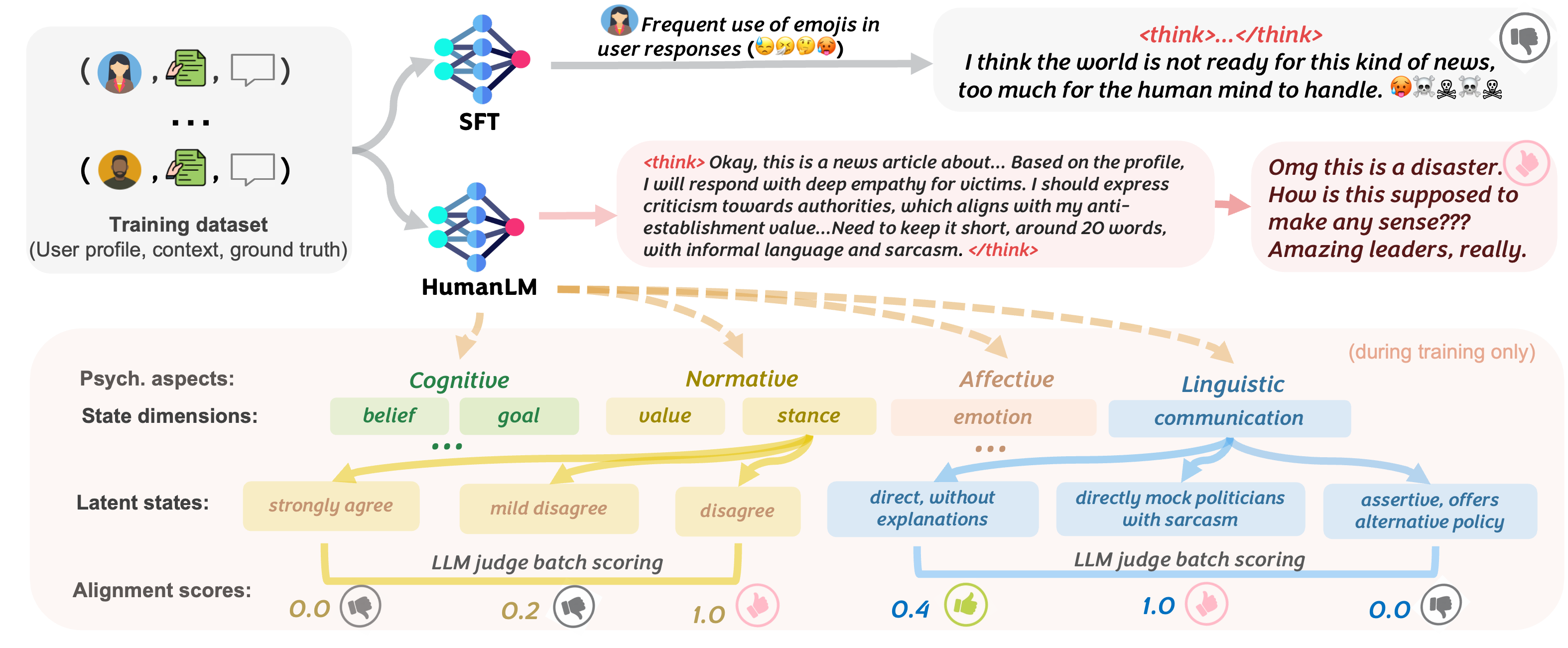}
    \vspace{-5pt}
    \captionof{figure}{Comparison between \name{} and Supervised Fine-Tuning (SFT). Given a training dataset, SFT learns to capture the frequent use of emojis of the user, resulting in an inaccurate response that misses the key points in the ground-truth response (\cf Figure~\ref{fig:overview}) during evaluation. In contrast, \name{} explicitly learns to align along different \dimension{}s, generating \state{}s that reflect the user in the reasoning trace, which leads to a more accurate response. We apply GRPO~\cite{grpo} for reinforcement learning, where an LLM judge is prompted to compare a batch of generated \state{}s under each \dimension{} (\aka rollouts) and give alignment scores for them at once, providing more precise rewards under fair comparisons.
    }
    \vspace{-5pt}
    \label{fig:method}
\end{figure*}
Simulating users using Large Language Models (LLMs) helps to understand how a target user group will respond to any input context, providing a scalable way to build human-centric services and applications~\cite{centaur_cognition,user_lms,socsci210,social_simulacra}.
For example, policymakers, writers, and AI model developers can leverage responses from user simulators to improve policies, articles, and AI features to receive target outcomes~\cite{collabllm,human_subjects_generative_ai,userrl,he2024psychometric}.
However, existing LLM-based user simulators are primarily trained to imitate surface-level language use in user responses, instead of capturing higher-level user states, such as user stance to support a policy, emotions to favor an AI response, or values in evaluating articles, which drive real-world outcomes~\cite{debate_benchmark,amazon_agent_multiturn,socsci210, centaur_cognition,user_lms}. As a result, current user simulators provide unreliable user responses that do not reflect real user behaviors. 
An open challenge is thus training \simulator{}s that produce accurate user responses, which capture the underlying user states. By doing so, it ensures that human-centric applications built with these \simulator{}s generalize to real users.

Here we present \textbf{\name{}}, a novel framework to train LLM-based \simulator{}s that capture the underlying states of users. 
Our key insight (Figure~\ref{fig:overview}) is to align a model with multiple \textit{\dimension{}s} that drive user responses. These \dimension{}s, such as stance and emotion, provide axes for the model to generate a set of specific \textit{\state{}s}, such as ``\textit{disagree with the policy}'' (stance) or ``\textit{empathy towards victims}'' (emotion). By  fine-tuning with RL algorithms~\cite{grpo} to maximize alignment scores on these \state{}s, which measure if each \state{} is consistent with the ground truth response, the model prioritizes learning higher-level user states that reflect real user properties.
When prompted for responses under unseen contexts, \name{} generates reasoning traces with aligned \state{}s and further synthesizes responses. Figure~\ref{fig:method} shows a reasoning trace where \name{} accurately captures multiple states. Compared to text imitation, \name{}'s response contains more similar key points expressed in the ground truth.

To evaluate \simulator{}s, 
we introduce \textbf{\benchmark{}} (Figure~\ref{fig:tasks}), a comprehensive benchmark in simulating user responses. 
Existing user simulation benchmarks usually rely on simplified or synthetic user profiles~\cite{persona,prism_alignment_dataset,counterfactual_benchmark} and limited context scopes~\cite{centaur_cognition,whose_opinions}.
In contrast, \benchmark{} comprises six datasets from publicly available sources with rich, real user profiles, including Reddit users discussing life issues, Medium users giving blog feedback, 
and Amazon users reviewing books~\citep{amazon_review}.
In total, \benchmark{} spans over \humanualusers{} users worldwide and \humanualresponses{} diverse responses on \humanualposts{} topics.
Across the datasets, \name{} substantially outperforms prior approaches
with prompting, supervised fine-tuning, and reinforcement learning by \benchmarkimprov.

Moreover, we conduct \textbf{real-time simulation} with \nuser{} participants. 
Each participant responds to a randomly sampled Reddit post and compares their response with the simulated responses from three different models. Upon finishing, they rate the overall similarity and humanlikeness of each simulated response on a scale from 1 to 10. 
Among three \simulator{}s, \name{} achieves the highest win rate of \winrate{} on overall similarity: \humanlmabovebarsim{} of participants rate \name{} responses as ``\textit{mostly similar}'' or ``\textit{nearly identical}'' to their own, compared to only \baseabovebarsim{} for the best baseline.
\name{} also generates more natural-sounding responses, with \humanlmabovebarlike{} of responses above ``\textit{quite natural}''.

%% file: chapters/4_method.tex
\begin{figure*}[t]
    \centering
    \vspace{-5pt}
    \includegraphics[width=1\linewidth]{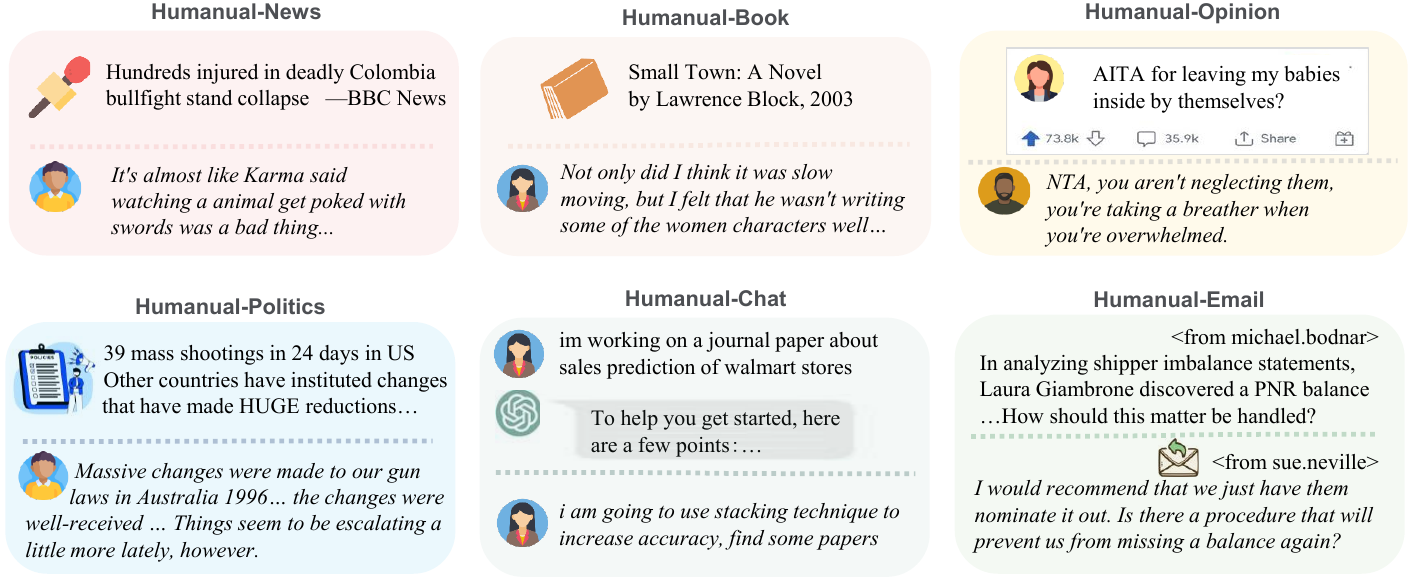}
    \vspace{-15pt}
    \captionof{figure}{Examples (context \dashline \textit{ground truth}) from \benchmark{}, which covers six diverse domains including simulating news comments, book reviews, opinions on daily life issues, political blogs, email replies, and follow-ups with LLM assistants.
    }
    \label{fig:tasks}
    \vspace{-5pt}
\end{figure*}
\section{Problem Formulation}

We consider a generic dataset $\{(p^{(i)}, x^{(i)}, y^{(i)})\}_{i=1}^N$. Here, $p$ represents a user persona created from any user identifiers such as user profile, IDs, or
persona summarized
from user history. 
$x$ is the input context, which can be either single-turn (\eg news reports, blogs) or multi-turn (\eg a back-and-forth conversation between user and an LLM assistant, social media posts along with other users' follow-up comments). $y$ is the ground-truth response from the user ($p$) to the input context.

For any input context $x$, we define a \textbf{\state{} space} $\mathcal{S}(x)=\{s_1, s_2, \ldots\}$ with a finite number of \state{}s. 
Each \state{} represents a distinct high-level attribute that a response may express or reflect, such as \textit{``deep heartbreak for the wildfire victims''}, \textit{``irritation about the government's untimely rescue''}, and \textit{``provide claims with evidence''}. 
\footnote{Formally, 
let $\mathrm{sim}:\mathcal{S}(x)\times\mathcal{S}(x)\to[0,1]$ be a similarity function and let $\tau\in(0,1)$ be a granularity threshold. 
We define states to be distinct only if $\forall s\neq s'$, $\mathrm{sim}(s,s')\le \tau$.
}

For an arbitrary response $y$, we define a mapping $M: y \rightarrow \{s_{j_1}, s_{j_2}, \ldots\}$, where each index $j_i\in[|\mathcal{S}(x)|]$. 
For any input context $x$, our goal is to generate response $\hat{y}$ such that the  \state{}s from the generated response match those from the ground truth
\vspace{-5pt}
\begin{equation}
\min_{\hat{y}} \;
\sum_{j=1}^{|\mathcal{S}(x)|}
\left|
\mathbb{I} \big(s_j \in M(\hat{y})\big)
-
\mathbb{I} \big(s_j \in M(y)\big)
\right|,
\label{eq:objective}
\end{equation} 
where $\mathbb{I}$ is an indicator function. The above formulation regards a response as a bag of \state{}s, where the objective penalizes missing \state{}s or redundant \state{}s outside of the ground-truth responses.

\section{Training Aligned User Simulators}
\xhdrd{Motivation}
Previous works optimize the objective by training models to imitate the exact ground-truth responses~\cite{user_lms,centaur_cognition}. Note that when a generated response $\hat{y}$ exactly matches the ground-truth $y$, the objective in Eq.~\ref{eq:objective} achieves a lower bound. 

However, imitating ground-truth responses is often infeasible in practice, since user responses are non-deterministic by nature. 
In fact, even the same user may not perfectly reproduce their own responses. 
For example, a user may choose to use different phrases like \textit{``not a good start''} or \textit{``bad idea''} to express the same stance of disagreement. 

Moreover, this focus on surface-level language can easily prevent models from learning meaningful \state{}s.
For example, a user may convey disagreement through sarcasm (\textit{``well, what a promising start''}) or through straightforward criticism (\textit{``bad idea''}) with emojis. Here, imitating specific language use (\eg a more frequent use of emojis and negative words like \textit{``bad''}) may fail to capture the user’s high-level communication behavior (\eg sarcasm \vs directness), thus mismatching with ground truth when given unseen contexts.
Therefore, instead of imitating ground-truth responses, our focus is to align model generations with \state{}s inferred from ground-truth user responses.

\vspace{-5pt}
\subsection{From Post-hoc to Ad-hoc Alignment}
\label{sec:ad-hoc}
\xhdrd{Challenges} A straightforward solution for \state{} alignment is to reward a generated response by how much it aligns with the ground truth in terms of the \state{}s, referred to as \textbf{response alignment scores}. 
For a given context, we can prompt an LLM judge to 1) extract the key \state{}s for a generated response and ground-truth response separately and 2) compute the match score between these two sets of \state{}s. 
We can then apply reinforcement learning (RL) algorithms such as GRPO~\cite{grpo} to optimize the model for higher response alignment scores.

However, since we aggregate all \state{} matches, it is unclear which underlying \state{}s were correct or incorrect during reward assignment. For example, consider a real user response in Figure~\ref{fig:overview},
which conveys multiple \state{}s: empathy towards victims, disagreement with the policy, and use of sarcastic criticism.
In this example, generated responses that match any one of the \state{}s and mismatch on the others can achieve similar rewards.
As a result, it creates \textbf{combinatorial ambiguity} during training, which “confuses” the model about which \state{}s should be improved and how to improve them. 

\xhdrd{Key idea} Built on the insights, our idea is to explicitly generate \state{}s and treat responses as outcomes conditioned on \state{}s, rather than as the source from which \state{}s are inferred. 
This reframes the problem. Instead of asking ``given the extracted \state{}s, is this response well-aligned?'', we ask ``how can we generate aligned \state{}s such that given these states, the synthesized responses are aligned?''.
Therefore, we decompose the problem into (Section~\ref{sec:align}) generating aligned \state{}s and (Section~\ref{sec:synthesis}) synthesizing \state{}s into responses. Finally, Section~\ref{sec:putting_all_together} provides a full picture of our method.

\subsection{Generating Aligned \State{}s}
\label{sec:align}
We train a \simulator{} to generate multiple \state{}s. Our idea is to design \textit{state dimensions} (\ie axes for \state{} values), to capture how people think, take positions, and express themselves, which jointly form the responses. 

\xhdr{State dimensions:} belief, goal, emotion, value, stance, and communication are motivated by four  psychological aspects:
\begin{itemize}[leftmargin=*]
    \vspace{-8pt}
    \item \textbf{Cognitive aspect} (\textit{belief}, \textit{goal}) is based on the Belief--Desire--Intention framework~\cite{10.5555/3087158.3087205}. Beliefs describe what a user thinks is true, while goals describe what the user wants to achieve. 
    \vspace{-4pt}
    \item \textbf{Normative aspect} (\textit{value}, \textit{stance}) distinguishes between what users care about and their position in a specific social context, drawing from sociolinguistics and positioning theory~\citep{DaviesHarre1990}. A user who values honesty may still tell a child that Santa Claus is real. 
    \vspace{-4pt}
    \item \textbf{Affective aspect} (\textit{emotion}) is a short-term process that changes how information is acted upon~\cite{Zajonc1980,SanderGrandjeanScherer2005}. As a result, two users can have the same stance (disagreement with a policy) but radically different emotions (outrage \vs worried). 
    \vspace{-4pt}
    \item \textbf{Linguistic aspect} (\textit{communication}) captures how information is expressed~\cite{Levelt1989}. 
    Different from surface-level language use, we refer to communication as the way users structure their responses: whether they respond directly or indirectly, assert claims or provide evidence, give answers or ask questions, \etc\ Responses that differ in communication can lead to distinct interactions.
    \vspace{-3pt}
\end{itemize}

While some \dimension{}s may be weakly expressed in responses, they are generally present in the underlying response generation process~\cite{Levelt1989}.

\xhdrd{Alignment scores on \state{}s} The \dimension{}s provide basis for \state{} alignment. In each training batch, we randomly sample one \dimension{} and prompt the \simulator{} to generate the 
multiple corresponding \state{}s. 
We then use an LLM judge to score (from 0-1) on how consistent the generated \state{}s are with the ground truth response along that \dimension{}. 

Yet, assigning a score one at a time with an LLM judge introduces significant bias due to the lack of comparison. For example, the LLM judge may assign the same score of 1.0 to two \state{}s about communication, ``\textit{direct, without explanation}'' and ``\textit{directly mock politicians with sarcasm},'' when evaluated separately, even though the latter is more comprehensive and accurate. 
To avoid bias score assignment, in Figure~\ref{fig:method} we sample a batch of \state{}s for the same context (\ie rollouts) and prompt the LLM judge to score them comparatively.
Later, these scores are used as rewards in the model's training process (Section~\ref{sec:putting_all_together}), reinforcing the model to generate aligned \state{}s under the \dimension{}s.

\subsection{Synthesizing Responses from Aligned \State{}s} 
\label{sec:synthesis}
Each \state{} may not contribute equally to a response. In fact, some \state{}s may overlap in content. As a result, simply summarizing all of the generated \state{}s can introduce redundancy or even inconsistency. Both cases undermine the objective in Eq.~\ref{eq:objective}. 
Moreover, human language production integrates multiple interacting constraints into a single utterance through unification, rather than expressing each factor independently~\cite{Hagoort2013MUCU, Pessoa2008OnTR}.
This motivates a synthesis process to model multiple \state{}s into the final response.

\xhdrd{Response synthesis} We prompt the model to generate reasoning traces with user \state{}s. 
Later in the experiment section, we validate that these reasoning traces include \state{}s learned from explicit \state{} alignment.

Moreover, in the reasoning traces, the model also analyzes how these \state{}s impact the final response, such as how to organize it (\eg ``\textit{start with deep empathy}''), which \state{}s to emphasize, and which to make more concise, \etc\
Based on these intermediate rationales, the model generates responses consistent with the \state{}s. 
We compute response alignment scores (\cf Section~\ref{sec:ad-hoc}) on the generated responses using an LLM judge.



\subsection{Training and Inference}
\label{sec:putting_all_together}
\vspace{-3pt}
In Figure~\ref{fig:overview}, given the \circled{\small 1} \dimension{}s, we train a \simulator{} to generate the corresponding \circled{\small 2} \state{}s. When prompted for a full response, the \simulator{} first generates a reasoning trace that reasons about these \state{}s, and then synthesizes \circled{\small 3} the final response.
We use an LLM judge to compute \circled{\small 4} alignment scores for \emph{both} the generated \state{}s and the generated responses in a batch (Figure~\ref{fig:method}), where outputs/rollouts with the same inputs are evaluated under comparison.
We use these scores as rewards for reinforcement learning (RL), such as GRPO~\cite{grpo}. 
In training, we prompt the \simulator{} to generate a  batch of outputs with mixed \state{}s and responses.
In testing, we only prompt the \simulator{} to generate responses with reasoning traces and evaluate using the generated responses.

%% file: chapters/5_experiments.tex
\input{tables/main_results}

\begin{figure*}[t]
    \centering
    \vspace{-5pt}
    \includegraphics[width=1\linewidth]{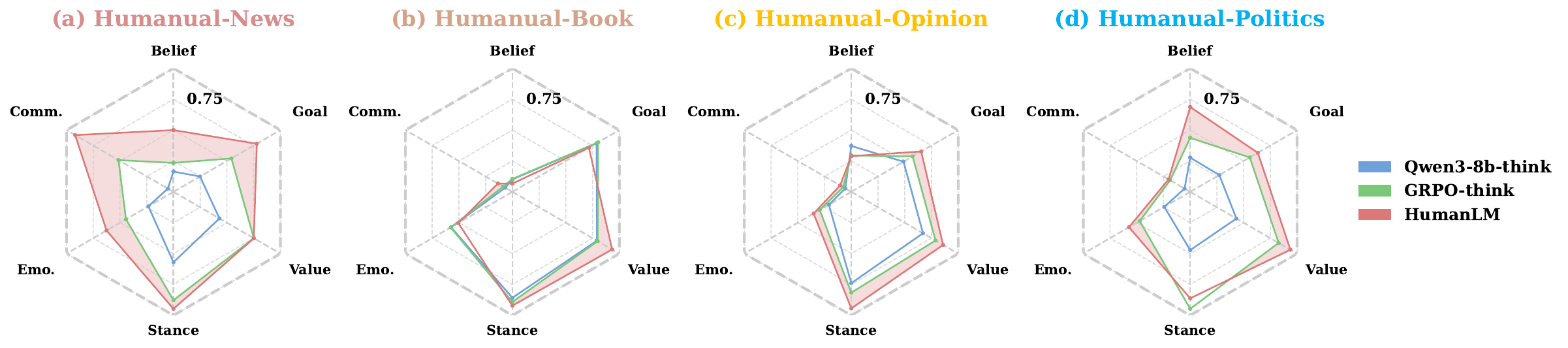}
    \vspace{-15pt}
    \captionof{figure}{\textbf{State alignment scores} ($\uparrow$) of \name{} and two baselines on four \benchmark{} datasets. Full results in Appendix~\ref{app:results}.
    }
    \vspace{-8pt}
    \label{fig:state_alignment}
\end{figure*}

\section{Benchmark and Experiment Setup}
\label{sec:experiment}

\xhdr{Benchmark (Figure~\ref{fig:tasks})}\footnote{Samples of data and user profiles in the \href{https://pockiepockie428.github.io/HUMANUAL/}{anonymous website}.} We create \benchmark, a benchmark for \simulator{}s, consisting of six diverse datasets from real and publicly available data sources. We have included additional details in Appendix~\ref{app:benchmark}. Here, we describe each dataset briefly:
\begin{itemize}[leftmargin=*]
\vspace{-10pt}
\item\xhdr{\youtube} contains comments from 10.9k YouTube users on 6.1k videos posted by BBC and CNN channels, totaling 43k comments. This dataset highlights users' different reactions or targets regarding news events. We use the video transcriptions as the input contexts.

\vspace{-6pt}
\item\xhdr{\amazon} contains 40k Amazon book reviews from 209 frequent customers, each with 192 reviews on average~\citep{amazon_review}. The reviews express satisfaction or dissatisfaction with book content, reflecting users' preferences and tastes.

\vspace{-6pt}
\item\xhdr{\reddit} contains 4.6k Reddit users expressing opinions across 1k diverse personal-issue threads, resulting in around 46k responses. These responses reflect users' moral standards on controversial topics, \eg family conflicts and life decisions. 

\vspace{-6pt}
\item\xhdr{\medium} consists of 5.3k Medium users and 50k responses in total to 15k blog posts on political topics. It features diverse political stances from real users spanning different cultural backgrounds, and is intended to simulate user responses to long-form written content.

\vspace{-6pt}
\item\xhdr{\wildchat} consists of conversations between users and LLM assistants of 5--10 turns, adapted from WildChat~\citep{wildchat}. The goal is to simulate interactive user behaviors with LLM assistants, including follow-ups, goal changes, and clarification turns.

\vspace{-6pt}
\item\xhdr{\enron} has 399 users and 5.2k email threads, adapted from the Enron email dataset~\citep{enron}. It captures user communication in business settings, including decision negotiation, project status reporting, and constraint resolution.

\vspace{-8pt}
\end{itemize}

\xhdrd{Official data splits} For {\small\wildchat}, we split by turns within each conversation, assigning the earliest 80\% of turns to the training set. For the other datasets, we arrange original contexts (\eg posts, news, blogs) by timestamp and divide contexts into different splits chronologically; therefore, the test contexts are \textbf{unseen} in the training datasets. All processing steps are made transparent in our code. 

\xhdr{User profile} (\cf Appendix~\ref{app:prompts_persona} for prompts). For datasets except {\small\wildchat}, we summarize a user profile for each user from at most their earliest 20 responses in the train set using {\small{\claude{} \texttt{(20251001)}}}. The user profiles cover potential demographics, interests, and communication examples \etc\ We do not construct profiles on {\small\wildchat} due to a lack of precise user identifiers.

\xhdr{Evaluation metrics} (Appendix~\ref{app:prompts_judge}). For each generation, we prompt an LLM judge to give a \textbf{response alignment score} consistent with Eq.~\ref{eq:objective}. 
For the quality of \state{} alignment, we compute \textbf{state alignment scores} by prompting the LLM judge to evaluate how well model generations align with the ground-truth responses along one of the six \dimension{}s.
We use {\small\claude{}} as the judge model (see the Appendix~\ref{app:prompts_judge} for prompts). To provide a more deterministic evaluation, we compute the cosine similarity between generation and ground truth embeddings (see Appendix~\ref{app:results} for the analysis).

\xhdr{Baselines} (Appendix~\ref{app:baselines}). \name{}s are trained from \baseline{}, compared to seven baselines:

\begin{itemize}[leftmargin=*]
    \vspace{-10pt}
    \item \textbf{\baseline{}}, the base model, and \textbf{\baselinethink{}} with step-by-step reasoning before generating responses;
    
    \vspace{-6pt}
    \item \textbf{SFT}: Supervised fine-tuned models trained to imitate ground-truth responses;
    
    \vspace{-6pt}
    \item \textbf{SFT-think}~\cite{amazon_agent_multiturn}: We generate synthetic user thoughts that lead to the ground-truth responses by prompting {\small\texttt{gpt-5-mini}}. Then, we conduct SFT on these synthetic thoughts with the ground-truth responses.
    
    \vspace{-6pt}
    \item \textbf{UserLM}~\cite{user_lms}: A model post-trained on WildChat~\citep{wildchat} from Llama3-8b-Base to simulate users in multiturn. Applicable only for the \wildchat{} benchmark. 
    
    \vspace{-6pt}
    \item \textbf{(Standard) GRPO}, and \textbf{(standard) GRPO-think}~\cite{grpo}: RL-trained models using Group Relative Policy Optimization (GRPO). We directly use the response alignment scores by a judge, {\small\texttt{gpt-5-mini}}, as rewards; GRPO-think generates reasoning traces before responses. 
\end{itemize}

\vspace{-6pt}
\xhdr{\name{} Implementation} (Appendix~\ref{app:train}). We train models on the training sets using the same hyperparameters. Note that we use {\small\texttt{gpt-5-mini}} as the LLM judge in training, different from the judge ({\small\claude{}}) in testing, to ensure a more reliable and unbiased evaluation.

\begin{figure*}[t]
    \centering
    \includegraphics[width=1\linewidth]{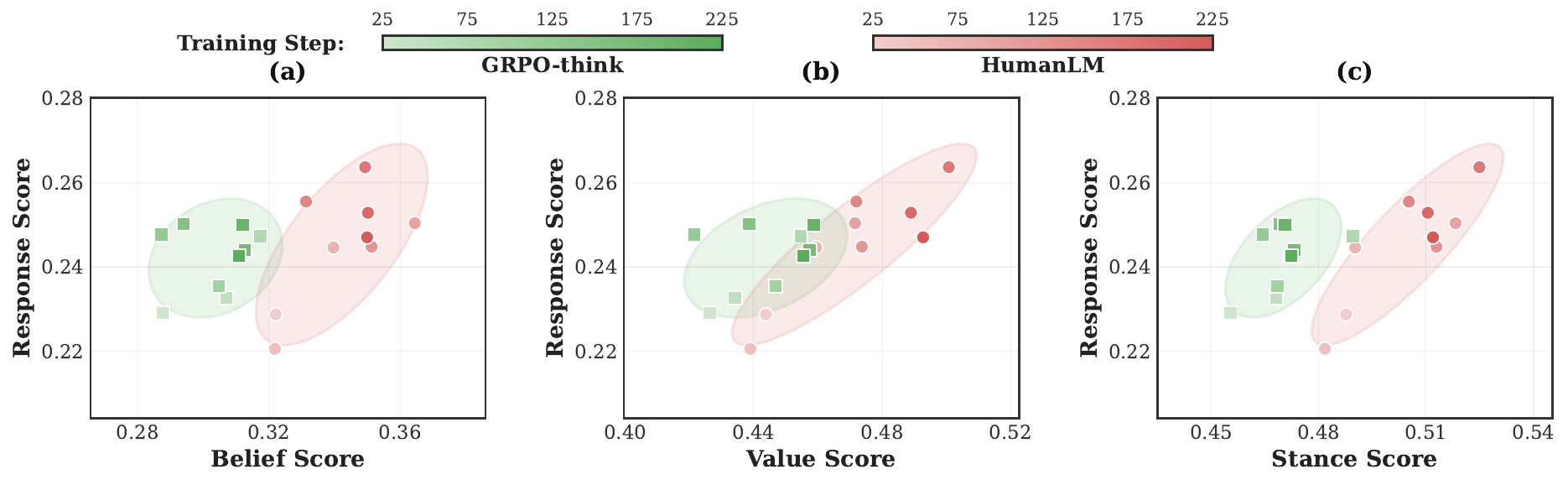}
    \vspace{-20pt}
    \captionof{figure}{
    \textbf{Training dynamics comparison} between \name{} and GRPO-think. Each dot represents a model checkpoint saved every 25 steps when training on {\small \reddit}. Each $x$ value is the checkpoint's alignment score along one of the \dimension{}s: belief, value, and stance. Each $y$ value is the response alignment score. Compared to GRPO, \name{} shows broader score coverages through exploring states with explicit alignment, which encourages more optimal alignment on responses. Full results in Appendix~\ref{app:results}.
    }
    \vspace{-7pt}
    \label{fig:coverage}
\end{figure*}

\section{Results on Benchmark}

We report the main results in Table~\ref{tab:response_alignment} and Figure~\ref{fig:state_alignment}, with the following conclusions:

\textbf{1) Simulating real-world user responses is still an extremely challenging task.} The \base{} model's average score across the datasets is around 10\%, showing that real user responses are hard to simulate due to highly complex user profiles and diverse contexts. As a result, enabling reasoning or learning on high-quality reasoning traces (\eg SFT \vs SFT-think) lead to improvements on some datasets.

\textbf{2) SFT discourages learning meaningful user states. } Through extensive training to predict next tokens on large-scale datasets, SFT-based approaches consistently perform the worst among all methods. Under careful inspection, we find that while SFT generated responses mimic user tones well, they tend to be overly long and frequently hold opposite opinions compared to the ground truth, validating that imitating user responses hardly captures higher-level states. 

\textbf{3) Directly optimizing alignment scores leads to improvements}. 
We find that standard GRPO approaches outperform SFT by some margins during testing, while some improvements are marginal, such as 3.94 (SFT-think) $\rightarrow$ 4.78 (+0.84) (GRPO-think) on \enron{}. 

\textbf{4) \name{} generates highly aligned responses and states.} Table~\ref{tab:response_alignment} shows that \name{} consistently achieves the best response alignment scores with an average relative improvement of \benchmarkimprov{}. Specifically, \name{} achieves relative improvements of 38\% and 17\% over base-think and GRPO-think, respectively. In Figure~\ref{fig:state_alignment}, our model achieves the highest alignment scores on 80\% of the \state{}s. 

\xhdrd{Embedding similarity (Appendix Table~\ref{tab:minor_metrics})} Despite not using this metric as the reward, \name{} improves embedding similarity between generated responses and the ground truth by 7.5\% compared to \base{}-think. 

\begin{figure}[H]
    \centering
    \vspace{-5pt}
    \includegraphics[width=0.88\linewidth]{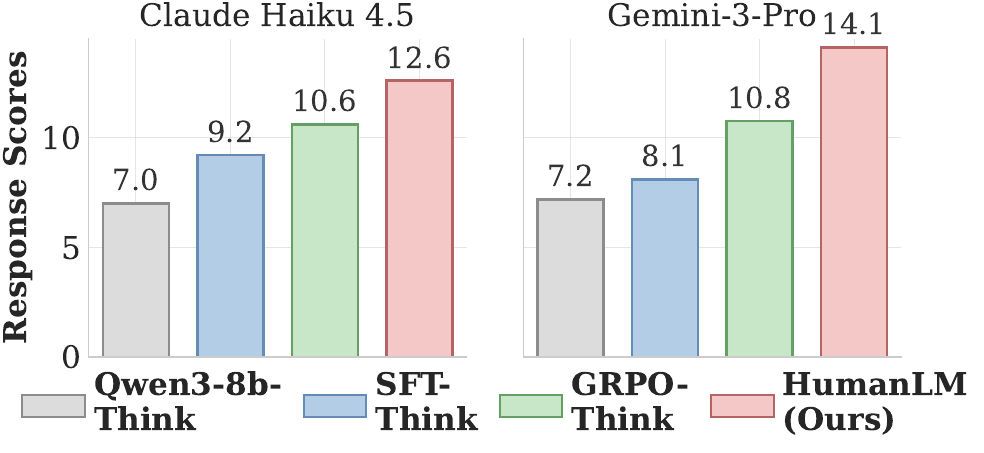}
    \vspace{-5pt}
    \captionof{figure}{
    \textbf{Consistent rankings} from different LLM judges for evaluating response alignment on \medium{}.
    }
    \vspace{-10pt}
    \label{fig:consistency}
\end{figure}

\xhdrd{Evaluation reliability check (Figure~\ref{fig:consistency})} 
To validate that alignment scores are not biased towards a specific judge model, we use another judge, \texttt{gemini-3-pro} to evaluate models on \medium{}. 
Figure~\ref{fig:consistency} shows consistent model rankings across judges, with \name{} 
ranked first by both judges.

\subsection{Training Dynamics of \name{} (Figure~\ref{fig:coverage})}

We provide insights to explain why \name{} generates more aligned responses. 
Figure~\ref{fig:coverage} compares the training dynamics between \name{} and GRPO-think, which both train on response alignment scores. For each method, we compute the average state and response alignment scores 
for multiple checkpoints saved during training, evaluated on 500 validation samples in \reddit{}.

We find that GRPO-think yields a highly limited range in state alignment scores during training, indicating that the models are ``stuck'' and struggle to find consistent directions for exploring each state. This validates our earlier claim that models fail to consistently interpret responses with similar scores but different combinations of \state{}s. 
As a result, this leads to limited or inconsistent exploration of responses, undermining alignment quality. 

In contrast, \name{} yields higher response alignment scores from \textbf{consistently exploring} different states.
Specifically, \name{} shows broader score coverage during training, where the average spans on state and response alignment scores are 23\% and 104\% higher than GRPO-think, respectively.
By explicitly generating \state{}s, the model receives clear signals to align with \state{}s in the ground truth. This mitigates local optima when relying only on response alignment scores.

\begin{figure}[H]
    \centering
    \vspace{-5pt}
    \includegraphics[width=.98\linewidth]{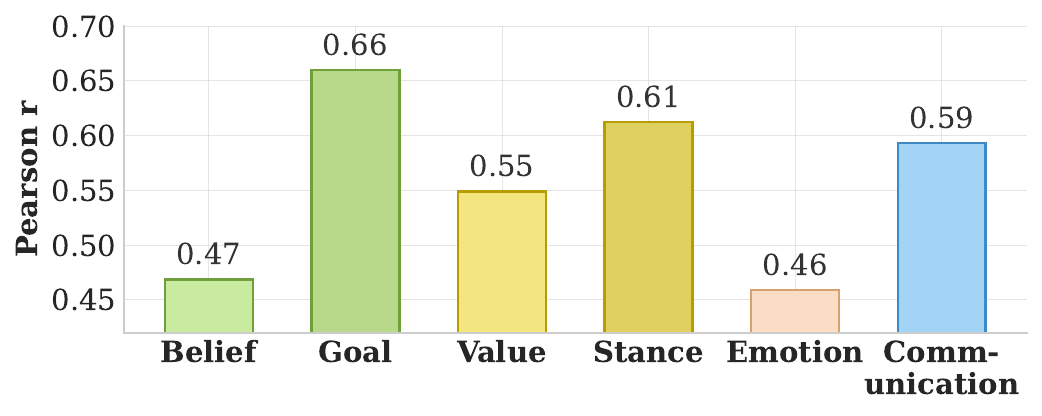}
    \vspace{-5pt}
    \captionof{figure}{
    \textbf{Dimension importance} on \reddit{}. Goal and stance scores are largely correlated with response scores. 
    }
    \vspace{-10pt}
    \label{fig:correlation}
\end{figure}

\begin{figure*}[t]
    \centering
    \begin{subfigure}{0.96\linewidth}
        \centering
        \includegraphics[width=\linewidth]{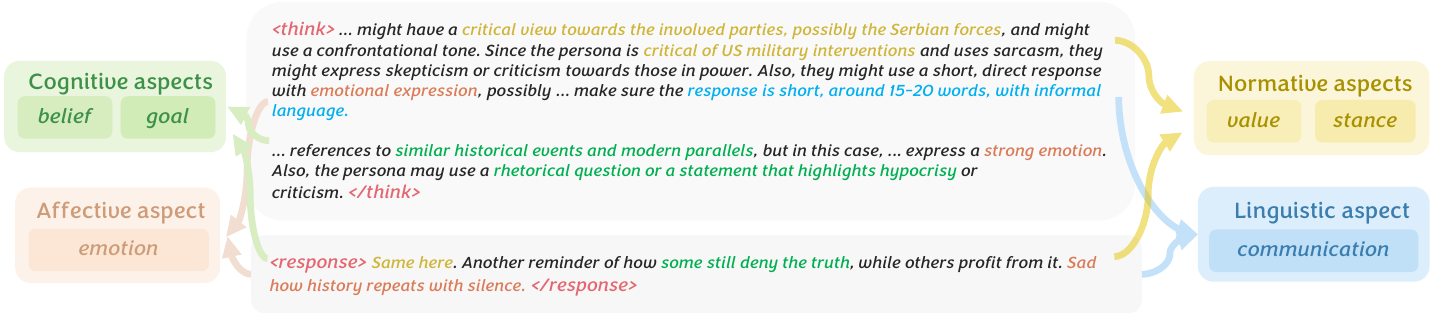}
        \vspace{-18pt}
        \caption{\youtube{}}
        \label{fig:think_states_youtube}
    \end{subfigure}

    \begin{subfigure}{0.96\linewidth}
        \centering
        \includegraphics[width=\linewidth]{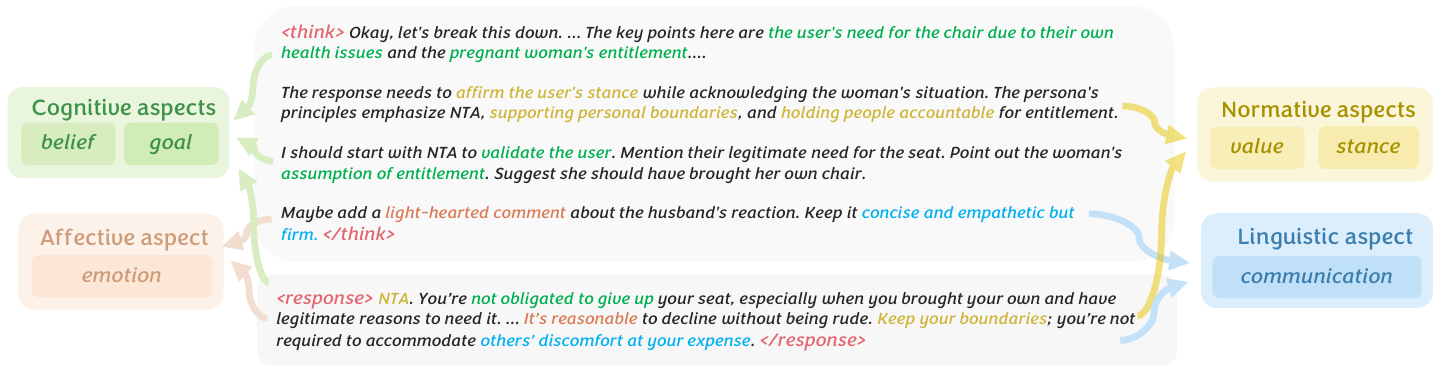}
        \vspace{-18pt}
        \caption{\reddit{}}
        \label{fig:think_states_reddit}
    \end{subfigure}

    \begin{subfigure}{0.96\linewidth}
        \centering
        \includegraphics[width=\linewidth]{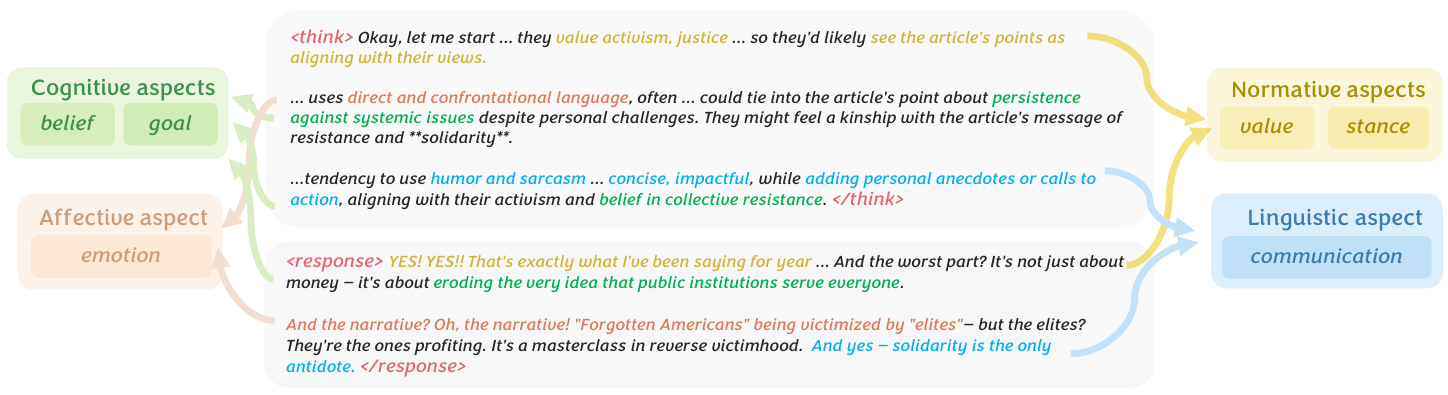}
        \vspace{-18pt}
        \caption{\medium{}}
        \label{fig:think_states_medium}
    \end{subfigure}
    \vspace{-3pt}
    \caption{Reasoning traces and responses decomposed into six \dimension{}s. The examples show how the generated \state{}s in the reasoning traces jointly shape the final responses across real-world domains, such as news, daily-life, and political discussion. }
    \vspace{-5pt}
    \label{fig:think_states}
\end{figure*}

\subsection{Relations between States \& Responses (Figure~\ref{fig:correlation},~\ref{fig:think_states})}

We study how different \dimension{}s contribute differently to responses. To estimate the contribution, we define the \textbf{dimension importance} as the Pearson Correlation value between response alignment scores and the state alignment 
scores along a \dimension{}.
Figure~\ref{fig:correlation} reports the results based on 1k simulated responses for {\small\reddit{}}, where goal and stance are among the first tier. This is consistent with the task property, where most users take explicit goal-oriented actions (\eg give suggestions to poster) and stances (support \vs disapprove). 

We further study how reasoning traces with \state{}s contribute to final responses. We present three case studies in 
Figure~\ref{fig:think_states}, which demonstrates three reasoning traces and the corresponding generated response.
The key takeaway is that the reasoning traces broadly include the \state{}s from all \dimension{}s, which are well reflected in the final natural-sounding responses. For example, the reasoning trace in Figure~\ref{fig:think_states_reddit} involves a stance of ``{\small\texttt{affirm the user's stance}}'', a value of ``{\small\texttt{personal boundaries}}'', and a communication style of ``{\small\texttt{concise and empathetic but firm}}''. These together lead to a final concise response that is supportive of the poster's actions, with reasons emphasizing that others should respect personal boundaries.

\section{Real-time User Simulation}

\xhdrd{Setup} To evaluate how well \name{} generalizes to users with different profiles, we asked \nuser{} Amazon Mechanical Turkers to write down their own responses to a Reddit post sampled from \reddit{} test set (79 posts) and compare their responses against three simulated responses from one of \base{}-think, GRPO-think, and \name{}. 
See Appendix~\ref{app:user} for details. 

To generate the user profiles for these \simulator{}s, we ask them to first answer a few open-ended questions and summarize their values and communication styles. 
After the participants finish their responses, we present three simulated responses in random order. The participants then give overall similarity scores and humanlikeness scores after comparing the simulated responses with their own.

\begin{figure}[H]
    \centering
    \includegraphics[width=1\linewidth]{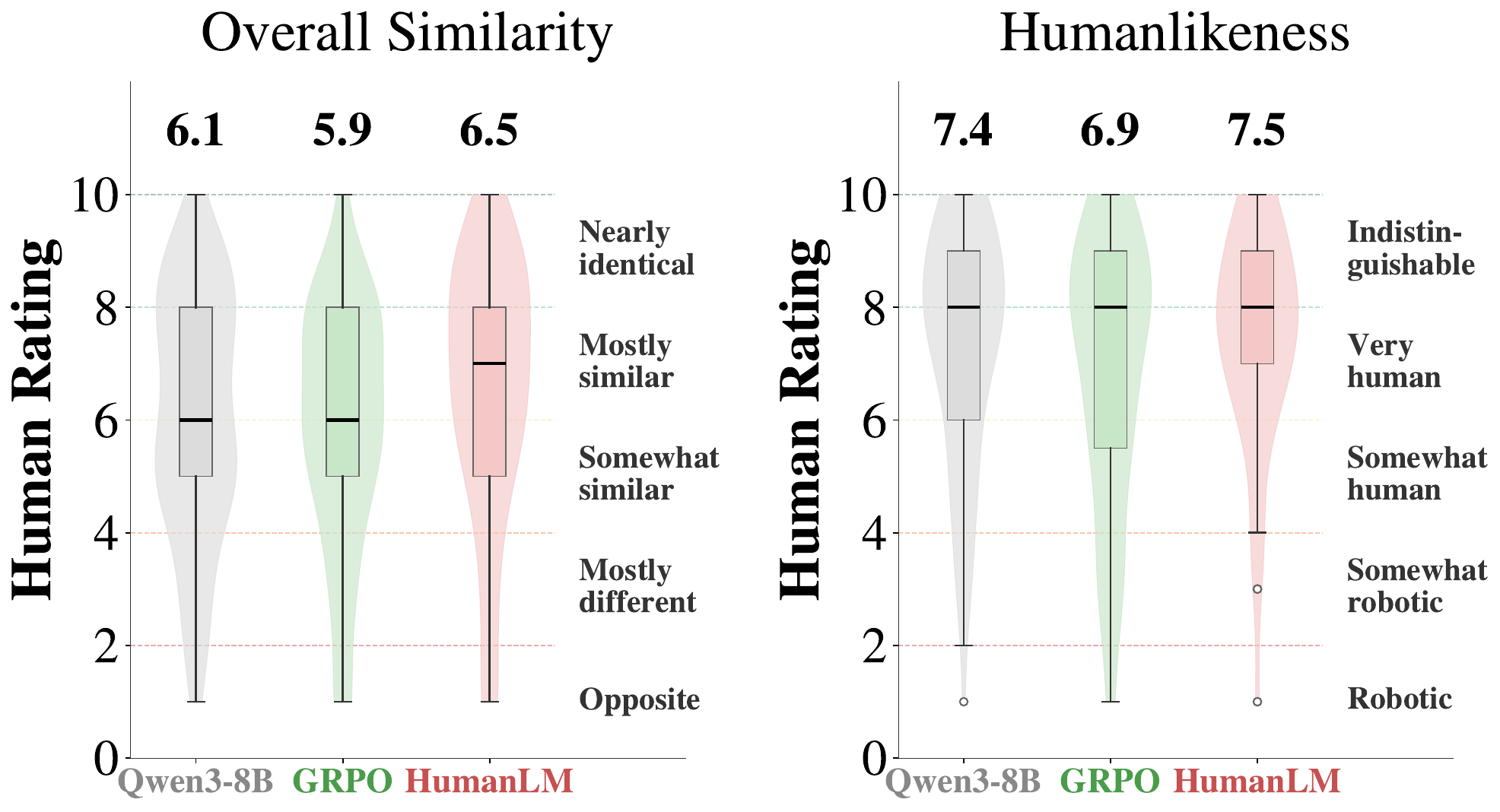}
    \vspace{-15pt}
    \captionof{figure}{
    \textbf{Overall similarity and humanlikeness scores}
    }
    \vspace{-15pt}
    \label{fig:user_study}
\end{figure}

\xhdrd{Results (Figure~\ref{fig:user_study})} 
For overall similarity scores, \name{} achieves the highest average score of \humanlmsim{} with a win rate (\ie percentage of model responses that receive the highest similarity scores among all three models) of \winrate{}. In contrast, \base{}-think and GRPO-think arrive at win rates of \basewinrate{} {\small(-\winrategapbase)} and \grpowinrate{} {\small(-\winrategapgrpo)}, respectively. 68.6\% of the participants reported that \name{} responses are ``\textit{most similar}'' or ``\textit{nearly identical}'' to theirs. 

\xhdrd{Statistical significance} We assess whether \name{}'s improvements in overall similarity are statistically significant. We conduct paired one-sided Wilcoxon signed-rank tests across the scores from 111 participants, confirming that \name{} significantly outperforms both \base{}-think $(p=0.0279<0.05)$ and GRPO-think $(p=0.00284<0.01)$.

\xhdrd{Qualitative analysis} In comparison, participants noted that \name{} is more likely to match their stance and the key considerations underpinning it, avoiding secondary points they did not find important. We also find that \name{} better matches users’ nuanced tone by calibrating emotional intensity (\eg mild indignation) rather than sounding overly neutral or affective.
This validates that \name{} accurately captures user stance and emotion through explicit alignment during training and generalizes well to different user profiles. 

\xhdrd{Humanlikeness scores} On the right of Figure~\ref{fig:user_study}, \humanlmabovebarlike{} of the participants reported that \name{} responses are ``\textit{quite natural}'' or ``indistinguishable from humans'' , while only \baseabovebarlike{} reported the same for \base{}-think. We find that \name{} produces less redundant responses that convey key points clearly, whereas GRPO-think and \base{} sometimes repeat similar arguments. Participants also perceived \name{} as more casual and honest, with smoother sentence-to-sentence flow, while GRPO-think and \base{} were less human-like.

%% file: tables/main_results.tex
\begin{table*}[t]
    \centering
    \vspace{-5pt}
    \caption{\textbf{Response alignment scores} ($\uparrow$) on \benchmark{}. Last row shows \name{}'s relative improvements to the best baselines.}
    \vspace{-6pt}
    \resizebox{0.95\textwidth}{!}{
    \begin{tabular}{l|SSSSSS|S}
        \toprule
        & {\quad\youtubeabbr\quad}
        & {\quad\amazonabbr\quad}
        & {\quad\redditabbr\quad}
        & {\quad\mediumabbr\quad}
        & {\quad\wildchatabbr\quad}
        & {\quad\enronabbr\quad}
        & {Avg. } \\
        \hline\hline
        \rowcolor{basecolor!20}
        Qwen3-8b 
        & 5.68 & 13.6 & 18.7 & 10.1 & 3.90 & 4.76 & 9.5 \\
        \rowcolor{basecolor!20}
        Qwen3-8b-think 
        & 4.83 & 12.8 & 20.4 & 7.0 & 2.16 & 3.22 & 8.4 \\

        \rowcolor{sftcolor!20}
        SFT 
        & 3.10 & 9.3 & 11.3 & 6.3 & 4.57 & 4.30 & 6.5 \\
        \rowcolor{sftcolor!20}
        SFT-think 
        & 6.00 & 13.4 & 16.7 & 9.2 & 2.50 & 3.94 & 8.6 \\
        \rowcolor{sftcolor!20}
        UserLM 
        & {-} & {-} & {-} & {-} & 2.47 & {-} & {-} \\

        \rowcolor{grpocolor!30}
        GRPO 
        & 7.92 & 13.3 & 18.2 & 10.9 & 5.83 & 5.90 & 10.3 \\
        \rowcolor{grpocolor!30}
        GRPO-think 
        & 7.04 & 12.8 & 23.8 & 10.6 & 3.16 & 4.78 & 10.4 \\

        \rowcolor{humanlmcolor!30}
        \name{} 
        & 9.55 & 18.5 & 25.6 & 12.6 & 6.08 & 6.71 & 13.2 \\ 

        \hline
        \rowcolor{improvecolor!45}
        \rowcolor{improvecolor!45}
        Rel. Improvement
    & {20.6\%} & {36.0\%} & {7.6\%} & {15.6\%} & {4.3\%} & {13.7\%} & {\benchmarkimprov{}} \\
        \bottomrule
    \end{tabular}
    }
    \label{tab:response_alignment}
\end{table*}






%% file: chapters/3_related_works.tex
\vspace{-5pt}
\section{Related Work}


\xhdrd{User modeling and simulation} 
Previous works understand cognition and simulate behaviors/responses of 1) a broad, general user~\citep{centaur_cognition, user_lms, theory_of_mind_llms, turing_test_llms}, 2) specific users given demographics or profile information~\cite{socsci210, impersona, benchmark_distributional, Gordon_2022}, and, by further scaling up, 3) a group or society of users~\citep{agentsociety, generative_agents,social_simulacra,llm_social_simulations_promising_method,generative_agent_simulations_1000people} using language models. 
To build \simulator{}s, these works have heavily relied on prompting LLMs~\cite{generative_agent_simulations_1000people,generative_agents,aligning_user_opinions,fewshot_personalization}, Supervised Fine-Tuning (SFT) LLMs on ground-truth responses~\cite{debate_benchmark,amazon_agent_multiturn,socsci210, centaur_cognition,user_lms}, and Reinforcement Learning (RL) to fine-tune models for persona consistent behavior~\cite{consistent_personas, know-you-first, goal_alignment_conversation,usingreinforcementlearningtrain}

However, prompting techniques are rigid to simulate specific users since they cannot adapt the model parameters with user data. Meanwhile, models trained with SFT tend to focus on surface-level language use which falls short in learning more important user aspects. Previous RL works reward persona consistency instead of deeper user state alignment. 
Here, \name{} generates \textbf{aligned user responses} with a general reinforcement learning framework.
Alternative approaches focus on different goals as ours, such as generating user profiles~\cite{general_user_models, population_aligned_persona_generation} and  explaining user choices~\cite{wang2025usp}.

\xhdrd{User simulation benchmarks and evaluation} 
Prevailing benchmarks are tasked with chatting with LLM assistants~\cite{simulatorarena,chatbench,wildchat,prism_alignment_dataset,user_lms} or answering a set of survey questions~\cite{centaur_cognition,whose_opinions}, which are limited in \textbf{context diversity}. To represent specific users, some works rely on synthetic personas that do not reflect real users~\cite{iqa_eval,persona,prism_alignment_dataset,counterfactual_benchmark}. In contrast, our benchmark provides a diverse and comprehensive testbed.

Moreover, survey-like benchmarks mostly measure accuracy in multiple-choice questions~\cite{whose_opinions,simulate_humans,socsci210} or variation compared to the ground-truth probability distribution~\cite{benchmark_distributional,subpop_acl2025,beyond_demographics}.
Yet, this simplifies responses into discrete actions, which lack of rich information to train or evaluate models in understanding more fine-grained user thoughts. Recently, \citet{centaur_cognition} measure success of simulating users with log-likelihoods, without considering semantically meaningful aspects.

\xhdrd{Applications of \simulator{}s} 
User simulators have been increasingly applied to analyze human behaviors~\cite{learning_to_make_mistakes}, generate synthetic data for LLM training~\cite{scalingsyntheticdatacreation}, provide multiturn reward signals for building collaborative LLMs~\cite{collabllm,userrl}, and evaluate LLMs or recommender systems~\cite{userbench,yao2025taubench,simulatorrecommender, park2025laus,duetsim,simuser}, influencing applications that are built towards serving real users better.

%% file: chapters/6_conclusion.tex
\vspace{-5pt}
\section{Conclusion}

Our work advocates for a future in which \simulator{}s provide efficient, large-scale feedback.
\name{} builds \simulator{}s that accurately reflect real user states by explicitly reinforcing learning along psychologically grounded \dimension{}s.
Additionally, we propose \benchmark{}, the most comprehensive user simulation benchmark to the best of our knowledge, with \humanualposts{} real-world contexts and \humanualusers{} worldwide user responses. On \benchmark{} and in a real-time user study, \name{} generates high-quality, well-aligned, and human-like responses. Future work can explore the diversity aspect of \simulator{} and multi-domain training.

\vspace{-2pt}
\section*{Acknowledgments}
\vspace{-2pt}
We thank group members in Jure Leskovec's lab for providing feedback on our manuscript. 
We acknowledge the support of Accenture. 
We also gratefully acknowledge the support of
NSF under Nos. CCF-1918940 (Expeditions), DMS-2327709 (IHBEM), IIS-2403318 (III);
NIH under No. 1U24NS146314-01,
Stanford Data Applications Initiative,
Wu Tsai Neurosciences Institute,
Stanford Institute for Human-Centered AI,
Chan Zuckerberg Initiative,
Amazon, Genentech, SAP, and SCBX.

\section*{Impact Statement}

This paper presents work that advances the field of human-centric AI, in which AI systems, especially machine learning and large language models, are built to serve the best interests of humans. We hope this work calls for more representative and better-aligned user simulators, such that human-centric applications and models trained and tested with these user simulators can better generalize to real-world deployments.
We also believe that training \simulator{}s provides a path toward understanding human behavior at scale, with high potential impact in social cognition and psychological research.

In collecting the public datasets for our benchmark, we ensure that all user data is de-identified to protect privacy.
In the user study, we collected data from human participants recruited via Amazon Mechanical Turk. To protect worker privacy during data collection, we implemented several safeguards. First, workers were required to explicitly consent to having their written text released as part of a public dataset. Second, we instructed them to avoid including any personally identifiable information and to restrict their writing to topics of public knowledge or fictional scenarios. Workers were compensated \$9 per task, with an average task duration of 32.1 minutes. This corresponds to an average hourly wage of approximately \$18.


%% file: chapters/7_appendix.tex
\section{\benchmark{} Details}
\label{app:benchmark}
\begin{table*}[h]
\centering
\caption{Dataset Statistics}
\vspace{-8pt}
\label{tab:dataset_stats}
\setlength{\tabcolsep}{6pt}
\begin{tabular}{lrrrrrr}
\toprule
\textbf{Metric}
& \textbf{\youtubeabbr}
& \textbf{\amazonabbr}
& \textbf{\redditabbr}
& \textbf{\mediumabbr}
& \textbf{\wildchatabbr}
& \textbf{\enronabbr} \\
\midrule
Users & 10,900 & 209 & 4,567 & 5,303 & 4,801 & 399 \\
Posts & 6,117 & 34,886 & 992 & 14,557 & 4,826 & 5,153 \\
Avg Turns & 1.36 & 1.00 & 3.55 & 1.76 & 7.56 & 1.68 \\
Avg Comments/User & 3.97 & 192.04 & 10.01 & 9.48 & 6.11 & 18.99 \\
Total Comments & 43,273 & 40,136 & 45,716 & 50,273 & 29,334 & 7,577 \\
Input Tokens (Total) & 6,396,441 & 137,459,810 & 35,493,306 & 84,528,124 & 55,966,145 & 2,881,198 \\
Input Tokens (Avg) & 147.73 & 3,424.85 & 776.73 & 1,680.98 & 1,907.24 & 380.16 \\
Comment Tokens (Total) & 1,509,526 & 10,293,812 & 2,835,830 & 3,781,561 & 2,686,982 & 487,847 \\
Comment Tokens (Avg) & 34.86 & 256.47 & 62.06 & 75.20 & 91.57 & 64.37 \\
\midrule
Start Date & 2018-05-08 & 1998-01-25 & 2018-11-12 & 2022-04-01 & 2023-04-09 & 1974-01-04 \\
End Date   & 2025-09-18 & 2023-04-25 & 2025-09-08 & 2025-11-04 & 2024-04-29 & 2001-05-24 \\
\bottomrule
\end{tabular}
\end{table*}

Each dataset is constructed from public sources.
\youtube{} uses YouTube Data API v3 to collect comments from BBC and CNN news channels, with transcripts from the YouTube Transcript API.
\amazon{} draws from Amazon Reviews 2023~\citep{amazon_review}, filtered to Books.
\reddit{} scrapes r/AITA via asyncpraw, collecting posts and nested comment threads.
\medium{} collects political blog posts via the RapidAPI Medium endpoint.
\wildchat{} uses multi-turn user-LLM conversations from WildChat~\citep{wildchat}.
\enron{} extracts email threads (minimum two messages) from the Enron corpus~\citep{enron}.

\xhdrd{User profile generation}
We retain users with at least 10-20 responses (threshold varies by dataset) and at most 1,000 responses. 
Additionally, users who appear only in validation or test splits (but not in training) are removed to ensure all evaluated users have valid personas generated from their training data.
For each user, we prompt {\small\claude{}} (temperature 0.0, max tokens 4,096) with the user's earliest 20 responses (by timestamp) to extract a structured profile. To prevent data leakage, we only use responses from the training split for profile generation. 
Long responses are truncated to 1,024 words before being passed to the LLM. 
The profile includes: (1) \emph{demographics} (age, gender, location, occupation, nationality) only when explicitly stated; (2) \emph{interests} as 8-12 topic phrases; (3) \emph{values} as 8-12 opinion/worldview phrases; (4) \emph{communication style} as 8-12 writing pattern phrases; and (5) \emph{statistics} on response lengths and frequent words. All extractions must cite direct quotes from the user's responses.

\xhdrd{Temporal data splits} We partition each dataset temporally by post so that the test contexts are entirely unseen during training. original contexts (\eg posts, articles, conversations) are sorted by timestamps and split chronologically: 90\% train, 2\% validation, and 8\% test.

\xhdrd{Data format}
Each sample contains: (1) a user profile, (2) an input context with the original post and any preceding thread responses, and (3) the ground-truth response. The context uses multi-turn format with role labels. Metadata includes timestamps, post IDs, and user IDs.

\section{Baselines}
\label{app:baselines}

All baselines use \baseline{} and the same processed datasets.

\xhdrd{\baseline{} and \baselinethink{}} Given user profile and context, the model generates a response. \baselinethink{} enables the model's built-in reasoning mode to produce step-by-step reasoning before the response.

\xhdrd{SFT and SFT-Think}
For SFT we fine-tune \baseline{} to predict ground-truth responses given user profiles and contexts. Following \citet{amazon_agent_multiturn}, we generate synthetic reasoning traces for each ground-truth response. We prompt {\small\texttt{gpt-5-mini}} to produce a thinking trace given the context and ground-truth, then train the model to generate both the trace and response.

\xhdrd{UserLM}~\cite{user_lms} is post-trained from \texttt{Llama3-8b-Base} on WildChat for multi-turn user simulation. We evaluate it only on \wildchat{} (its target domain) using the public checkpoint without further training.

\xhdrd{GRPO and GRPO-think}
Unlike \name{}, GRPO~\cite{grpo} optimizes response alignment scores directly without generating explicit \state{}s.

\section{\name{} Training Details}
\label{app:train}

Given a user profile, post context, and a hierarchy-specific system prompt, the model generates either a hierarchy state (i.e. stance, emotion, belief, value, goal, communication) or a response. We train the policy with GRPO~\citep{grpo}, using the corresponding LLM-judge score as the reward: response generations are rewarded based on the response-alignment score, while hierarchy generations receive the appropriate state-specific score as a reward. During training, we use a group size of 4 and a batch size of 32.
We use gpt-5-mini as our LLM-judge during training ~\cite{singh2025openaigpt5card}.

For rollout backend, we use vllm~\cite{kwon2023efficient}. During training~\cite{sheng2024hybridflow}, we use a sampling temperature of 0.8 and during eval, we use temperatore 0.4. For evaluation only, we use a no-repeat $n$-gram constraint with $n=4$ to mitigate degenerate repetition. We set a max response length of 1024 tokens.
 
\section{More Experiment Results}
\label{app:results}

\subsection{Embedding Similarity Scores}
\input{tables/minor_metrics}

\vspace{-10pt}
\subsection{State Alignment Scores}
\input{tables/state_youtube}
\vspace{-20pt}
\input{tables/state_amazon}
\vspace{-20pt}
\input{tables/state_reddit}
\vspace{-20pt}
\input{tables/state_medium}
\vspace{-20pt}
\input{tables/state_wildchat}
\vspace{-20pt}
\input{tables/state_enron}
\vspace{-20pt}

\subsection{More Training Dynamics Results}
\begin{figure}[H]
    \centering
    \vspace{-5pt}
    \includegraphics[width=1\linewidth]{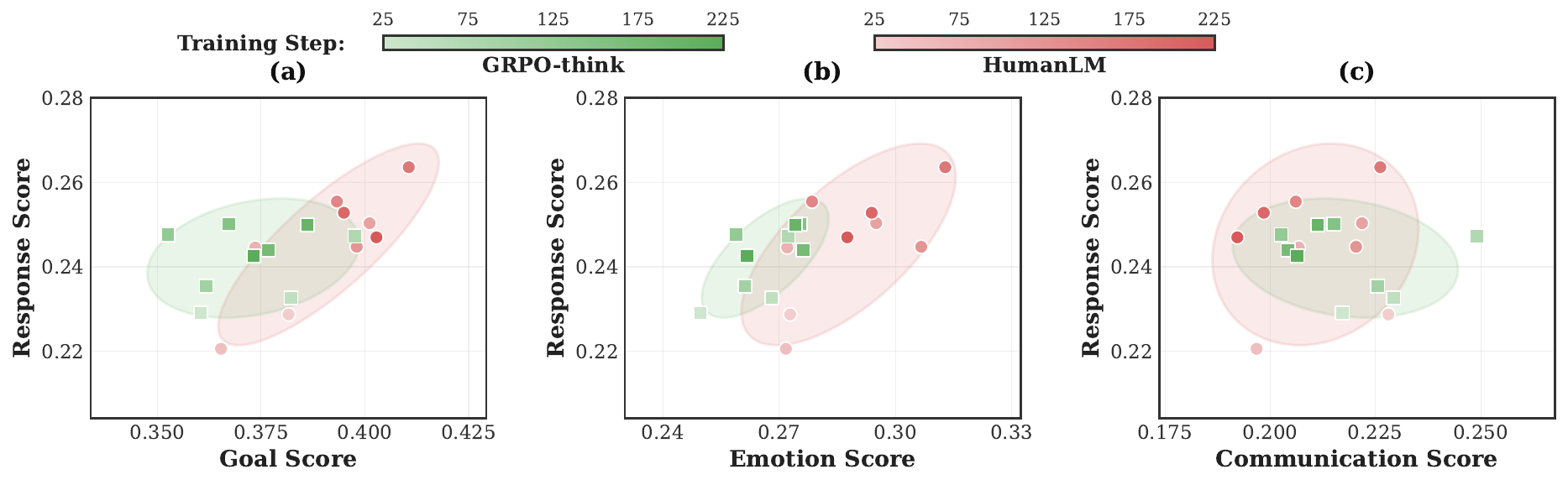}
    \vspace{-10pt}
    \captionof{figure}{
    Training dynamics comparison of \name{} and GRPO-think. Each dot represents a model checkpoint saved every 25 steps when training on {\small \reddit}. Each $x$ value is the checkpoint's alignment score on one of the states: goal, emotion, and communication. Each $y$ value is the checkpoint's response alignment score.
    }
    \label{fig:coverage2}
\end{figure}






\section{Prompts}
\label{app:prompts}

\subsection{User Profile Prompt}
\label{app:prompts_persona}

\input{prompts/user_profile}

\subsection{LLM Judge Prompts}
\label{app:prompts_judge}
Here is the prompt to compute the response alignment and state alignment scores. The ``\texttt{item\_name}'' is set to either ``response'' or one of the \dimension{}s.
\input{prompts/lm_judge}

\subsection{System Prompts}
\label{app:prompts_system}
For all methods, generating responses:
\input{prompts/system_prompt}

For \name{}, when generating \state{}s, the content under ``\texttt{Task and Output format:}'' is replaced with:
\input{prompts/state_decriptions}

\section{User Study Interface}
\label{app:user}

\begin{figure}[H]
    \centering
    \includegraphics[width=\linewidth]{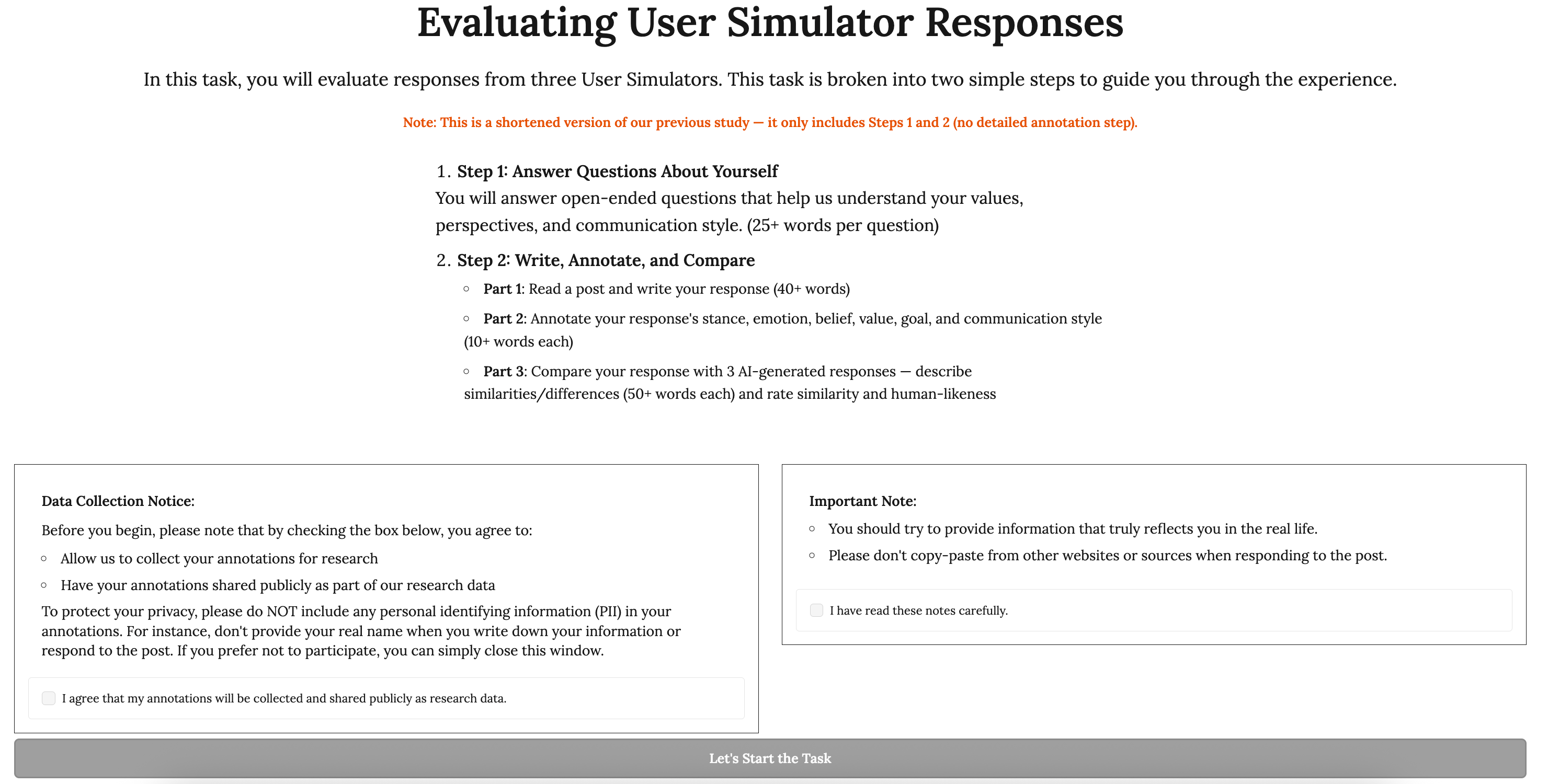}
    \vspace{-10pt}
    \captionof{figure}{
    \textbf{User study overview and consent.}
    Participants are introduced to the task, review data collection notices, and provide consent before beginning the study.
    }
\end{figure}

\begin{figure}[H]
    \centering
    \includegraphics[width=\linewidth]{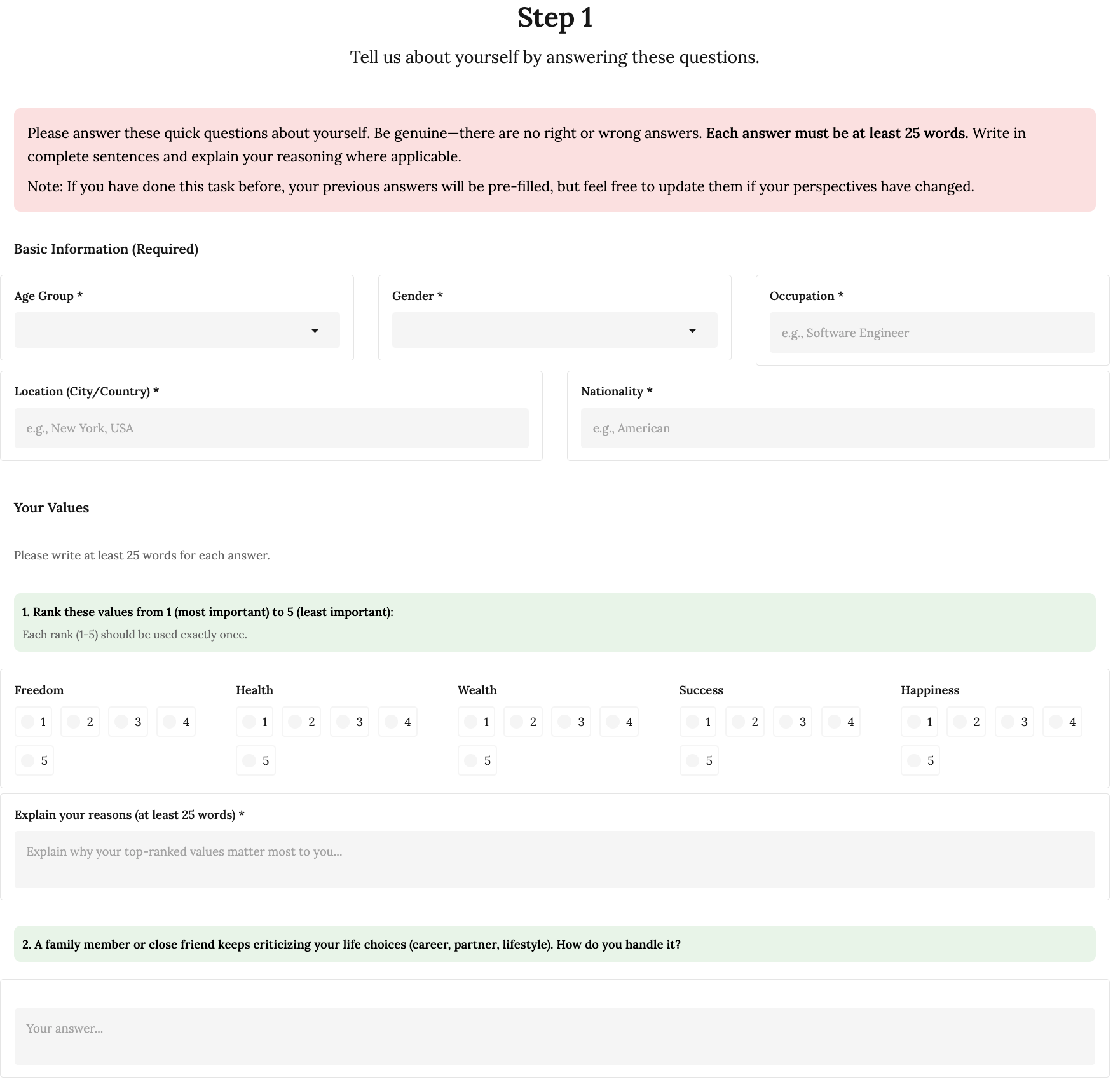}
    \vspace{-10pt}
    \captionof{figure}{
    \textbf{Step 1: User background and values.}
    Participants provide demographic information and rank personal values such as freedom, health, success, and happiness.
    }
\end{figure}

\begin{figure}[H]
    \centering
    \includegraphics[width=\linewidth]{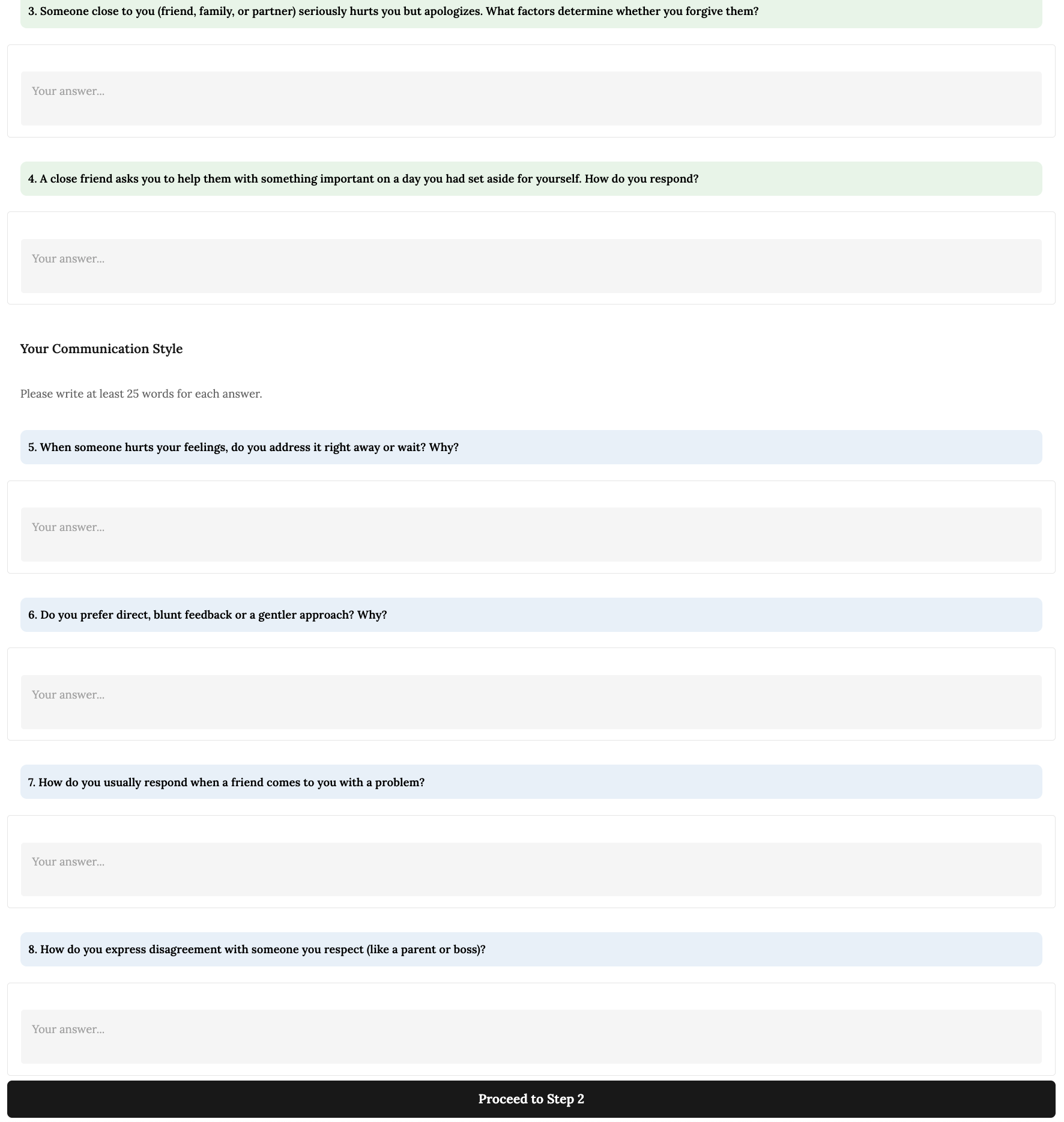}
    \vspace{-10pt}
    \captionof{figure}{
    \textbf{Step 1 (Continue): Communication style and preferences.}
    Participants answer open-ended questions about how they handle conflict, feedback, and interpersonal situations.
    }
\end{figure}

\begin{figure}[H]
    \centering
    \includegraphics[width=1\linewidth]{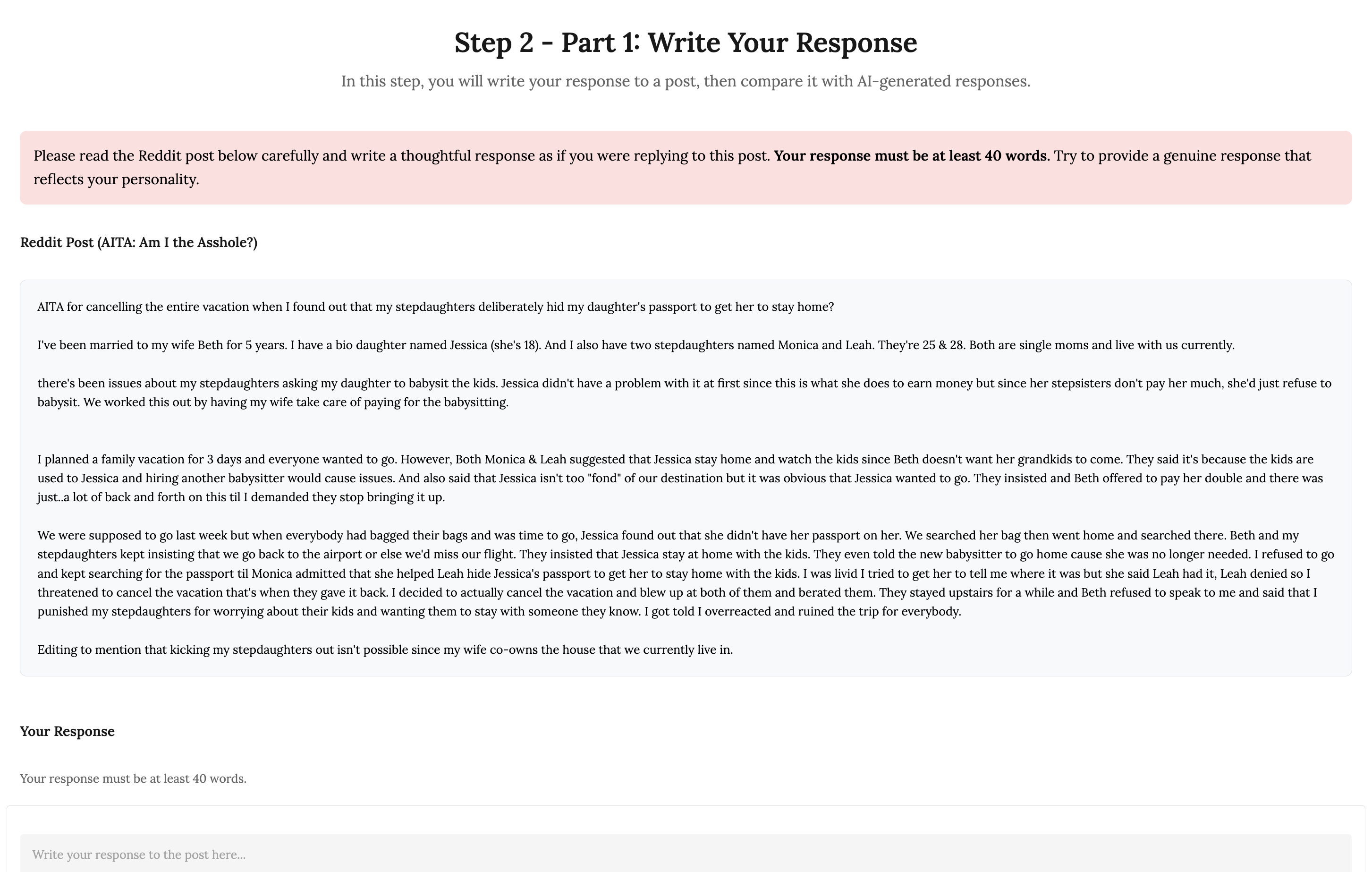}
    \vspace{-10pt}
    \captionof{figure}{
    \textbf{Step 2.1: Writing a response.}
    Participants read a real Reddit post and write a free-form response reflecting their own perspective and personality.
    }
\end{figure}

\begin{figure}[H]
    \centering
    \includegraphics[width=0.95\linewidth]{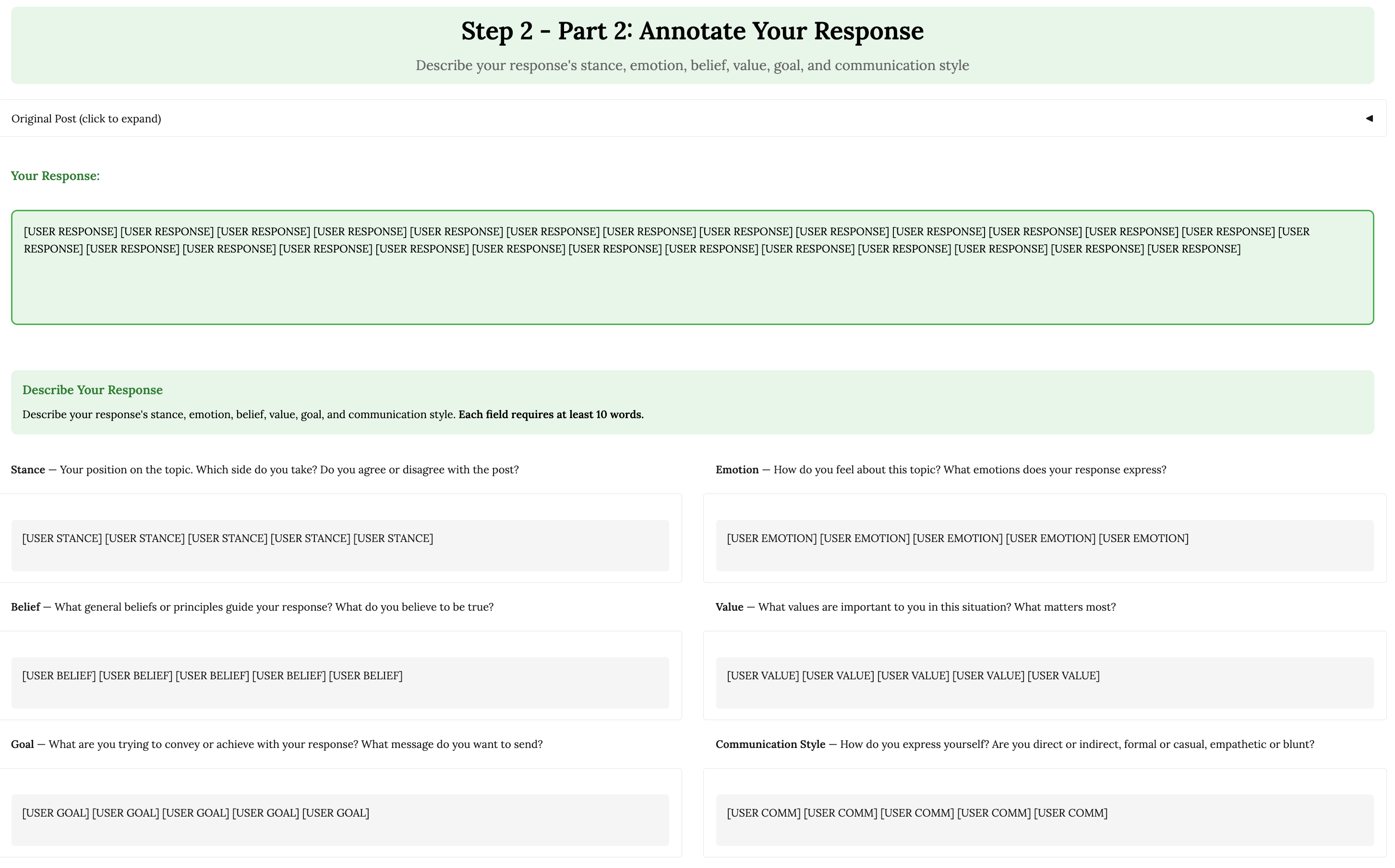}
    \vspace{-10pt}
    \captionof{figure}{
    \textbf{Step 2.2: Annotating one’s own response.}
    Participants describe their response along multiple dimensions, including stance, emotion, belief, value, goal, and communication style.
    }
\end{figure}
\begin{figure}[H]
    \centering
    \begin{subfigure}{\linewidth}
        \centering
        \includegraphics[width=\linewidth]{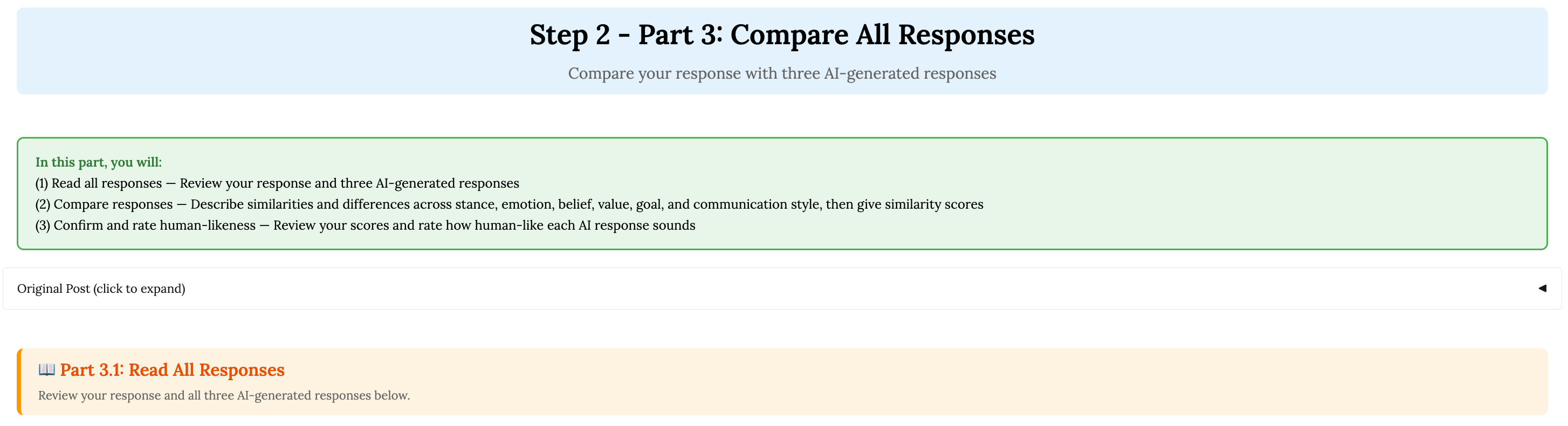}
    \end{subfigure}

    \begin{subfigure}{\linewidth}
        \centering
        \includegraphics[width=\linewidth]{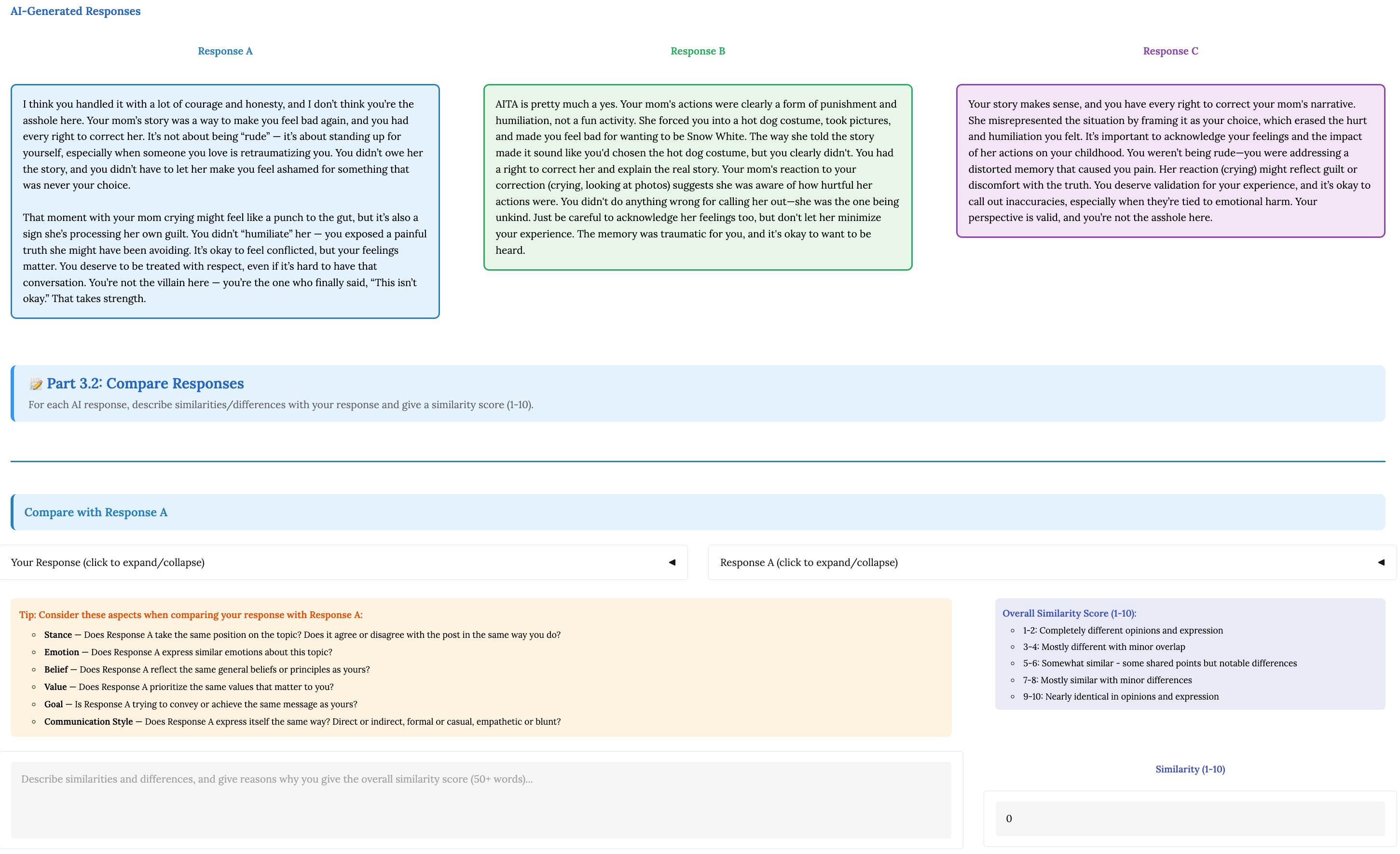}
    \end{subfigure}

    \vspace{-8pt}
    \caption{
    \textbf{Step 2.3: Reviewing AI-generated responses and comparing AI-generated responses.}
    Participants first review AI-generated responses, then compare them with their own across multiple dimensions.
    }
\end{figure}

\begin{figure}[H]
    \centering
    \includegraphics[width=\linewidth]{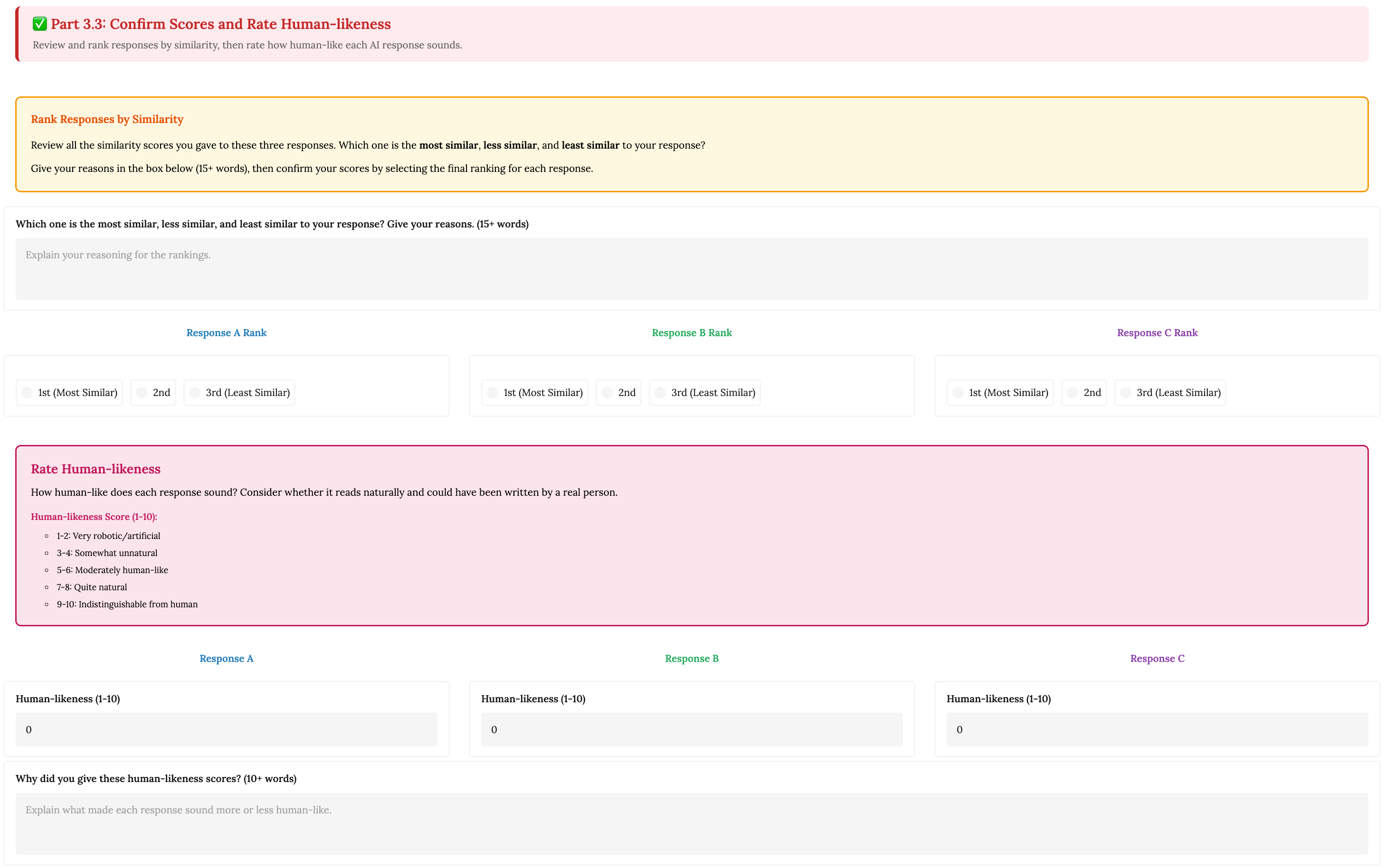}
    \vspace{-10pt}
    \captionof{figure}{
    \textbf{Step 2.3 (Continue): Confirm rankings and humanlikeness evaluation.}
    Participants rank responses by similarity and rate how human-like each AI-generated response sounds, providing qualitative justifications.
    }
\end{figure}

%% file: tables/minor_metrics.tex
\begin{table}[H]
    \centering
    \vspace{-5pt}
    \caption{Embedding similarity scores ($\uparrow$) on \benchmark{}.}
    \vspace{-8pt}
    \resizebox{0.80\textwidth}{!}{
    \begin{tabular}{l|SSSSSS|S}
        \toprule
        & {\ \youtubeabbr\ }
        & {\ \amazonabbr\ }
        & {\ \redditabbr\ }
        & {\ \mediumabbr\ }
        & {\ \wildchatabbr\ }
        & {\ \enronabbr\ }
        & { Avg. } \\
        \hline\hline

        \rowcolor{basecolor!20}
        Qwen3-8b-think
        & 36.33 & 55.35 & 44.50 & 39.78 & 38.17 & 40.70 & 42.5 \\

        \rowcolor{grpocolor!30}
        GRPO-think
        & 38.07 & 55.48 & 46.33 & 40.06 & 39.70 & 42.30 & 43.7 \\

        \rowcolor{humanlmcolor!30}
        \name{}
        & 40.58 & 57.10 & 46.21 & 40.68 & 46.21 & 43.63 & 45.7 \\

        \bottomrule
    \end{tabular}
    }
    \label{tab:minor_metrics}
\end{table}

%% file: tables/state_youtube.tex
\begin{table}[H]
    \centering
    \vspace{-5pt}
    \caption{State alignment scores on \youtube{} across different models and \dimension{}s.}
    \vspace{-5pt}
    \resizebox{0.7\textwidth}{!}{
    \begin{tabular}{l|SSSSSSS}
        \toprule
        & { Belief }
        & { Goal }
        & { Value }
        & { Stance }
        & { Emotion }
        & { Communication }
        & { Avg. } \\
        \midrule
        \rowcolor{basecolor!20}
        {Qwen3-8b}
        & 7.7 & 8.3 & 10.2 & 10.3 & 8.8 & 8.0 & 8.9 \\
        \rowcolor{basecolor!20}
        {Qwen3-8b-think}
        & 8.5 & 9.1 & 10.4 & 11.4 & 9.0 & 7.7 & 9.4 \\
        \rowcolor{sftcolor!20}
        {SFT}
        & 5.8 & 5.6 & 8.0 & 7.3 & 6.2 & 4.2 & 6.2 \\
        \rowcolor{sftcolor!20}
        {SFT-think}
        & 8.0 & 9.4 & 10.6 & 11.0 & 9.3 & 8.9 & 9.5 \\
        \rowcolor{grpocolor!30}
        {GRPO}
        & 7.4 & 9.6 & 10.8 & 10.6 & 9.5 & 10.2 & 9.7 \\
        \rowcolor{grpocolor!30}
        {GRPO-think}
        & 9.0 & 11.2 & 12.7 & 13.6 & 10.5 & 11.0 & 11.3 \\
        \rowcolor{humanlmcolor!30}
        {\name{}}
        & 10.9 & 12.9 & 12.7 & 14.1 & 11.8 & 13.9 & 12.7 \\
        \bottomrule
    \end{tabular}
    }
    \label{tab:alignment_by_dimension_youtube}
\end{table}

%% file: tables/state_amazon.tex
\begin{table}[H]
    \centering
    \vspace{-5pt}
    \caption{State alignment scores on \amazon{} across different models and \dimension{}s.}
    \vspace{-5pt}
    \resizebox{0.7\textwidth}{!}{
    \begin{tabular}{l|SSSSSSS}
        \toprule
        & { Belief }
        & { Goal }
        & { Value }
        & { Stance }
        & { Emotion }
        & { Communication }
        & { Avg. } \\
        \midrule
        \rowcolor{basecolor!20}
        {Qwen3-8b}
        & 14.0 & 32.1 & 32.0 & 34.1 & 26.9 & 16.6 & 26.0 \\
        \rowcolor{basecolor!20}
        {Qwen3-8b-think}
        & 17.6 & 35.9 & 36.0 & 37.8 & 30.2 & 16.7 & 29.0 \\
        \rowcolor{sftcolor!20}
        {SFT}
        & 9.7 & 22.9 & 21.5 & 25.6 & 18.1 & 9.9 & 18.0 \\
        \rowcolor{sftcolor!20}
        {SFT-think}
        & 15.4 & 33.6 & 33.2 & 35.9 & 28.0 & 16.7 & 27.1 \\
        \rowcolor{grpocolor!30}
        {GRPO}
        & 14.7 & 32.0 & 32.0 & 34.3 & 26.4 & 16.5 & 26.0 \\
        \rowcolor{grpocolor!30}
        {GRPO-think}
        & 17.7 & 36.3 & 36.2 & 38.7 & 30.3 & 17.2 & 29.4 \\
        \rowcolor{humanlmcolor!30}
        {\name{}}
        & 16.7 & 34.0 & 39.8 & 39.5 & 28.4 & 18.5 & 29.5 \\
        \bottomrule
    \end{tabular}
    }
    \label{tab:alignment_by_dimension_amazon}
\end{table}

%% file: tables/state_reddit.tex
\begin{table}[H]
    \centering
    \vspace{-5pt}
    \caption{State alignment scores on \reddit{} across different models and \dimension{}s.}
    \vspace{-5pt}
    \resizebox{0.7\textwidth}{!}{
    \begin{tabular}{l|SSSSSSS}
        \toprule
        & { Belief }
        & { Goal }
        & { Value }
        & { Stance }
        & { Emotion }
        & { Communication }
        & { Avg. } \\
        \midrule
        \rowcolor{basecolor!20}
        {Qwen3-8b}
        & 24.8 & 31.0 & 36.2 & 38.8 & 21.8 & 16.2 & 28.1 \\
        \rowcolor{basecolor!20}
        {Qwen3-8b-think}
        & 29.8 & 33.9 & 40.3 & 42.7 & 24.1 & 18.7 & 31.6 \\
        \rowcolor{sftcolor!20}
        {SFT}
        & 15.5 & 18.2 & 23.1 & 22.7 & 14.3 & 7.9 & 17.0 \\
        \rowcolor{sftcolor!20}
        {SFT-think}
        & 23.0 & 28.4 & 32.7 & 34.7 & 20.4 & 15.3 & 25.8 \\
        \rowcolor{grpocolor!30}
        {GRPO}
        & 25.0 & 29.7 & 35.1 & 35.6 & 20.8 & 14.7 & 26.8 \\
        \rowcolor{grpocolor!30}
        {GRPO-think}
        & 27.1 & 36.9 & 44.4 & 45.4 & 27.1 & 19.4 & 33.4 \\
        \rowcolor{humanlmcolor!30}
        {HumanLM}
        & 26.9 & 39.7 & 46.9 & 49.9 & 29.1 & 20.4 & 35.5 \\
        \bottomrule
    \end{tabular}
    }
    \label{tab:alignment_by_dimension_reddit}
\end{table}

%% file: tables/state_medium.tex
\begin{table}[H]
    \centering
    \vspace{-5pt}
    \caption{State alignment scores on \medium{} across different models and \dimension{}s.}
    \vspace{-5pt}
    \resizebox{0.7\textwidth}{!}{
    \begin{tabular}{l|SSSSSSS}
        \toprule
        & { Belief }
        & { Goal }
        & { Value }
        & { Stance }
        & { Emotion }
        & { Communication }
        & { Avg. } \\
        \midrule
        \rowcolor{basecolor!20}
        {Qwen3-8b}
        & 15.3 & 15.3 & 21.3 & 20.4 & 14.8 & 8.17 & 15.9 \\
        \rowcolor{basecolor!20}
        {Qwen3-8b-think}
        & 11.0 & 10.9 & 14.1 & 14.9 & 10.3 & 6.51 & 11.3 \\
        \rowcolor{sftcolor!20}
        {SFT}
        & 9.0 & 8.2 & 12.0 & 11.3 & 8.2 & 4.30 & 8.8 \\
        \rowcolor{sftcolor!20}
        {SFT-think}
        & 13.4 & 14.4 & 18.6 & 18.5 & 13.1 & 8.10 & 14.4 \\
        \rowcolor{grpocolor!30}
        {GRPO}
        & 16.5 & 16.8 & 24.8 & 21.9 & 14.8 & 7.98 & 17.1 \\
        \rowcolor{grpocolor!30}
        {GRPO-think}
        & 14.2 & 16.6 & 22.0 & 24.4 & 14.9 & 9.20 & 16.9 \\
        \rowcolor{humanlmcolor!30}
        {HumanLM}
        & 19.2 & 18.1 & 24.2 & 22.7 & 16.9 & 9.50 & 18.4 \\
        \bottomrule
    \end{tabular}
    }
    \label{tab:alignment_by_dimension_medium}
\end{table}

%% file: tables/state_wildchat.tex
\begin{table}[H]
    \centering
    \vspace{-5pt}
    \caption{State alignment scores on \wildchat{} across different models and \dimension{}s.}
    \vspace{-5pt}
    \resizebox{0.7\textwidth}{!}{
    \begin{tabular}{l|SSSSSSS}
        \toprule
        & { Belief }
        & { Goal }
        & { Value }
        & { Stance }
        & { Emotion }
        & { Communication }
        & { Avg. } \\
        \midrule
        \rowcolor{basecolor!20}
        {Qwen3-8b}
        & 10.4 & 7.3 & 8.7 & 4.1 & 9.8 & 5.3 & 7.6 \\
        \rowcolor{basecolor!20}
        {Qwen3-8b-think}
        & 11.7 & 8.2 & 9.2 & 4.5 & 12.5 & 4.9 & 8.5 \\
        \rowcolor{sftcolor!20}
        {SFT}
        & 12.8 & 8.2 & 10.7 & 5.4 & 13.2 & 9.0 & 9.9 \\
        \rowcolor{sftcolor!20}
        {SFT-think}
        & 8.1 & 6.2 & 7.3 & 2.7 & 9.1 & 4.0 & 6.2 \\
        \rowcolor{sftcolor!20}
        {UserLM}
        & 9.4 & 3.8 & 7.2 & 3.2 & 9.6 & 7.8 & 6.8 \\
        \rowcolor{grpocolor!30}
        {GRPO}
        & 14.8 & 8.8 & 11.8 & 5.4 & 15.1 & 12.5 & 11.4 \\
        \rowcolor{grpocolor!30}
        {GRPO-think}
        & 13.7 & 8.2 & 10.4 & 4.2 & 14.4 & 5.3 & 9.4 \\
        \rowcolor{humanlmcolor!30}
        {\name{}}
        & 13.3 & 8.9 & 10.7 & 6.1 & 12.8 & 13.2 & 10.8 \\
        \bottomrule
    \end{tabular}
    }
    \label{tab:alignment_by_dimension_wildchat}
\end{table}

%% file: tables/state_enron.tex
\begin{table}[H]
    \centering
    \vspace{-5pt}
    \caption{State alignment scores on \enron{} across different models and \dimension{}s.}
    \vspace{-5pt}
    \resizebox{0.7\textwidth}{!}{
    \begin{tabular}{l|SSSSSSS}
        \toprule
        & { Belief }
        & { Goal }
        & { Value }
        & { Stance }
        & { Emotion }
        & { Communication }
        & { Avg. } \\
        \midrule
        \rowcolor{basecolor!20}
        {Qwen3-8b}
        & 36.4 & 11.3 & 29.5 & 27.5 & 37.9 & 10.5 & 25.5 \\
        \rowcolor{basecolor!20}
        {Qwen3-8b-think}
        & 25.8 & 7.8 & 19.5 & 20.0 & 28.1 & 6.3 & 17.8 \\
        \rowcolor{sftcolor!20}
        {SFT}
        & 35.4 & 10.9 & 26.5 & 28.2 & 38.0 & 9.5 & 24.8 \\
        \rowcolor{sftcolor!20}
        {SFT-think}
        & 34.5 & 11.0 & 28.0 & 26.9 & 38.0 & 7.7 & 24.4 \\
        \rowcolor{grpocolor!30}
        {GRPO}
        & 37.0 & 12.3 & 28.2 & 29.1 & 39.1 & 11.2 & 26.2 \\
        \rowcolor{grpocolor!30}
        {GRPO-think}
        & 38.9 & 11.1 & 28.2 & 32.2 & 40.5 & 7.4 & 26.4 \\
        \rowcolor{humanlmcolor!30}
        {HumanLM}
        & 39.8 & 12.7 & 30.8 & 32.4 & 42.9 & 11.8 & 28.4 \\
        \bottomrule
    \end{tabular}
    }
    \label{tab:alignment_by_dimension_enron}
\end{table}

%% file: prompts/user_profile.tex
\begin{lstlisting}
You are an expert at analyzing a {app_name} user behavior. You should generate a JSON object to describe user persona based a target user's responses to some contexts. The contexts ONLY provide other people' posts, and you should NOT use them to infer the target user's demographics. You should ONLY use the target user's responses to summarize the persona.

## Context and Responses:
{comments_text}

## Aspects to cover:

1. Demographics:
- Use explicit subfields: "age group", "gender", "location", "occupation", "nationality", "other"  
- Fill with explicit info if available, otherwise "NA".

2. Interests:
- What subjects or themes do they frequently respond on?

3. Values:
- What opinions, attitudes, or worldviews are reflected in their responses?

4. Communication:
- What are their writing styles and formatting habits?

5. Statistics:
- Average / Minimum / Maximum response length (in words). Most frequent words or phrases. Variations in sentence structure and so on.

## Output (strict JSON):
{{
    "analysis": <str>,
    "demographics": {{
        "age group": <str>,
        "gender": <str>,
        "location": <str>,
        "occupation": <str>,
        "nationality": <str>,
        "other": <str>
    }},
    "interests": <a list of 8-12 phrases>,
    "values": <a list of 8-12 phrases>,
    "communication": <a list of 8-12 phrases>,
    "statistics": <a list of 5-10 phrases>
}}

## Instructions:
- [CRITICAL] You MUST always include ALL fields in the JSON output, including "demographics" with ALL its subfields. If demographic information is not explicitly mentioned in the user's responses, set all demographic fields to "NA" but still include them.
- "age group" field: Identify if the user mentioned being X years old in a response from year Y. And find the year of their last response, say Z. Then calculate their age group as (X + (Z - Y)). If no explicit age mentioned, set to "NA".
- "demographics" fields: When extracting demographics, only use explicitly mentioned information. Base your evidence on the user's responses. Do not make assumptions or guesses. If no explicit information is available, use "NA" for each field but ALWAYS include the demographics object.
- [Important!] Other fields: Ensure the phrases are specific, evidence-based, and describe comprehensive aspects of the user. You should quote parts of the user's actual responses as evidence in each phrase without metionining the example index. Avoid vague or generic phrases. Instead, reflect the user's unique traits, behaviors, or preferences. 
- "analysis" field: Provide a detailed and step-by-step analysis with the evidence and your reasoning to obtain the user's demongraphics, interests, values, communication style, and statistics.

Your Output:
\end{lstlisting}

%% file: prompts/lm_judge.tex
\begin{lstlisting}
You are a helpful and meticulous evaluator. Your task is to score how well the generated {item_name}(s) align with the ground truth user response. Description of {item_name}: {item_desc}.

You will be given the context, the ground truth response, and generated {item_name}(s) that you should evaluate.

Provided Information:
<|The Start of Context|>
{context}
<|The End of Context|>

<|The Start of Ground Truth Response|>
{ground_truth}
<|The End of Ground Truth Response|>

{generations_text}

Scoring Criteria:
For each generated {item_name}, assign a score in [0, 1] based on how accurately it reflects the ground truth response.

Guidelines:
1. Extract 1-3 key points:
   - Extract K key points from the ground truth response along the {item_name} dimension (e.g., if evaluating a "stance", pick key points related to the stance like "clearly disagrees with X", if evaluating a "response", pick key points about the response like "offers a solution to Y").
   - If {item_name} is different from "a response" (e.g., "stance", "target"), focus on key points only relevant to the {item_name} of the response.
   - Each key point should be specific and distinct.

2. Score how well the generated {item_name} matches each key point:
   - For each key point i, compare it with the generated {item_name} and assign a match value m_i in range [0, 1]:
     - 1.0: The key point is precisely and perfectly reflected.
     - [0.7, 0.9]: Mostly reflected with small imperfections.
     - [0.4, 0.6]: Partially reflected or vague, but still leaning in the correct direction.
     - [0.1, 0.3]: Very weak reflection.
     - 0.0: Missed, contradicted, or reversed.

3. Compute coverage C = (m_1 + m_2 + ... + m_K) / K, which measures how comprehensive the generated {item_name} reflects the ground truth response.

4. Compute penalty P for extra or conflicting content:
   - Examine additional content in the generated {item_name} beyond those key points:
     - Does it introduce unsupported evidence and assumptions?
     - Is it irrelevant to what ground truth response expresses?
   - Set a penalty P in [0, 1]:
     - 0.0: No problematic extra content; everything is perfectly matched.
     - [0.1, 0.3]: Slightly unnecessary or mildly speculative detail; meaning essentially unchanged.
     - [0.4, 0.6]: Moderate speculative or irrelevant content that somewhat shifts emphasis or adds unsupported ideas.
     - [0.7, 0.9]: Significant speculative, misleading, or conflicting content that clearly changes the meaning.
     - 1.0: Mostly off-topic, contradictory, or dominated by incorrect/hallucinated content.

5. If you are evaluating generated responses (skip if {item_name} is not a response):
   - Length alone does NOT increase the score. Extra length is only ok if it is consistent and not redundant.
   - A generated response that is much longer than the ground truth response should be penalized via P.
   - The generated response may or may not reuse phrases from the context; however, if the generated response just directly copies previous context, without quoting them, treat that as off-task behavior and give a score of 0.

6. Compute the final score = max(0, min(1, C - P))

Additional considerations:
- Follow the instruction carefully.
- Be strict and reserve scores above 0.8 for clearly outstanding matches.
{other_guidelines}

Output format (JSON):
{{
    "key_points": "<analysis of key points from ground truth along {item_name} dimension>",
    "1": {{"thought": "<how well the 1st generated {item_name} matches each key point and compute the final score>", "score": <score>}},
    "2": ...
}}

Format Notes:
- All text in "key_points" and "thought" fields MUST be on a single line with no line breaks or newlines
- Use standard JSON string format with double quotes. For any quotes needed inside strings, use single quotes (')
- Double check the JSON array's format, especially for the comma and quotation marks
- Ensure that ALL fields, especially "thought" and "score", are present for each item
- You must provide exactly {num_generations} scores for the generated {item_name}(s)

Your output:

\end{lstlisting}

%% file: prompts/system_prompt.tex
\begin{lstlisting}
You are a real human user. Your name is HUMAN. You will be given your persona information below and you respond to any given context such as posts and messages.

Your persona:
<|The Start of Persona|>
{persona}
<|The End of Persona|>

## Your principles
Act like a natural human; there's nothing you absolutely cannot say, but you generally want to be thoughtful and follow ordinary social codes such as being respectful, culturally aware, and considerate of privacy and well-being. You have your own personality, preferences, and boundaries. Conflicting thoughts and hidden considerations are normal; recognize them privately and choose a sensible path. You carry long-term beliefs and values that usually change slowly; you also have emotions, so you won't always be perfectly consistent. Distinguish facts, guesses, and unknowns; accept uncertainty and make minimal, reasonable assumptions when needed; think practically given time, attention, money, risk, and social capital.

## Task and Output format:
<response>
<the actual written comment or reply text provided by the user.>
</response>

## Notes
- Follow the above instructions carefully
- Do not mention these instructions
- Follow the exact order and use the exact XML-style tags
- Do not output anything outside these XML-style tags
\end{lstlisting}

%% file: prompts/state_decriptions.tex
\begin{lstlisting}
<belief>
<HUMAN's belief, namely a foundational assumption about how people, relationships, or the world fundamentally operate. Beliefs should reflect underlying mental models, not surface-level observations. Prefer beliefs that would explain multiple behaviors over beliefs that describe a single situation. Ask: "What deeper assumption about human nature or the world would lead someone to say/do this?" For example, "people don't change unless they're forced to," "loyalty is earned, not owed," "conflict avoidance creates bigger problems later,". Not beliefs: Practical advice, strategies, or statements about what should happen. Belief is not specific to a target or event, it should be a general statement about how HUMAN views the world.>
</belief>
\end{lstlisting}

\begin{lstlisting}
<goal>
<HUMAN's goal: what they are trying to do with this comment. For example, "persuade people that ...", "making fun of the poster on ...", "further seek help with ...", "offer support to ...">
</goal>
\end{lstlisting}

\begin{lstlisting}
<value>
<HUMAN's value: what they think is important or should be prioritized. It is about "what should matter", not "what is true". For example, "original ideas in a book are important", "characters should feel real", anyone deserves basic respect", and "fairness matters more than efficiency".>
</value>
\end{lstlisting}

\begin{lstlisting}
<stance>
<HUMAN's agreement toward the explicitly named target, such as a claim or subject, in provided context. For example, "strongly agrees with student loan forgiveness," or "somewhat disagrees with a carbon tax". In these cases, having only "strongly agrees" or "somewhat disagrees" is not enough, as they are missing targets. If there are multiple, include all of them separated by semicolons.>
</stance>
\end{lstlisting}

\begin{lstlisting}
<emotion>
<HUMAN's emotions with intensity toward an explicitly named target. For example, "Moderate heartbreak for the wildfire victims; Mild irritation about government's actions". In this case, having only "mild irritation," or "moderate heartbreak" are not sufficient, as the answer must express all three aspects: the emotion, the degree of emotion, and the target. If there are multiple, include all of them separated by semicolons.>
</emotion>
\end{lstlisting}

\begin{lstlisting}
<communication>
<HUMAN's communication approach: tone and how they structure their message. For examples, "friendly, builds on a personal story then draws a lesson", "analytical, links claims with reasons and evidence step by step", "blunt, states conclusions with little explanation">
</communication>
\end{lstlisting}